\documentclass[12pt]{article}
\usepackage{graphicx,psfrag,epsf}
\usepackage{enumerate}
\usepackage{natbib}
\usepackage{url} % not crucial - just used below for the URL 
\usepackage{xcolor} % remember to comment out later
\usepackage{amsthm,amsmath}
\usepackage{amssymb,amsfonts}
\usepackage{graphicx,psfrag,epsf}
\usepackage{mdframed}
\usepackage{glossaries}
\RequirePackage{bbm}
\RequirePackage{booktabs,array,hhline,bm,tcolorbox,mathtools}
\usepackage{placeins,float}
\usepackage{comment}
\usepackage{multirow}

\usepackage[colorlinks=true, linkcolor=blue,citecolor=cyan]{hyperref}

\newcommand{\blind}{0}

\begin{document}

\def\spacingset#1{\renewcommand{\baselinestretch}%
{#1}\small\normalsize} \spacingset{1}

\def\B0{\mbox{\boldmath $0$}}
\def\Ba{\mbox{\boldmath $A$}}
\def\Bb{\mbox{\boldmath $B$}}
\def\Bc{\mbox{\boldmath $C$}}
\def\Bd{\mbox{\boldmath $D$}}
\def\Bl{\mbox{\boldmath $L$}}
\def\Bn{\mbox{\boldmath $N$}}
\def\Br{\mbox{\boldmath $R$}}
\def\Bt{\mbox{\boldmath $T$}}
\def\Bv{\mbox{\boldmath $V$}}
\def\Bx{\mbox{\boldmath $X$}}
\def\By{\mbox{\boldmath $Y$}}
\def\Bz{\mbox{\boldmath $Z$}}

\def\btheta{\mbox{\boldmath $\theta$}}
\def\htheta{\mbox{$\hat{\theta}$}}

\def\balpha{\mbox{\boldmath $\alpha$}}
\def\bbeta{\mbox{\boldmath $\beta$}}
\def\bdelta{\mbox{\boldmath $\delta$}}
\def\bepsilon{\mbox{\boldmath $\epsilon$}}
\def\bGamma{\mbox{\boldmath $\Gamma$}}
\def\blambda{\mbox{\boldmath $\lambda$}}
\def\bLambda{\mbox{\boldmath $\Lambda$}}
\def\bmu{\mbox{\boldmath $\mu$}}
\def\bSigma{\mbox{\boldmath $\Sigma$}}
\def\btau{\mbox{\boldmath $\tau$}}
\def\btheta{\mbox{\boldmath $\theta$}}

\definecolor{steelblue}{RGB}{70, 130, 180}
\definecolor{darkorange}{RGB}{255, 140, 0}
\definecolor{salmon}{RGB}{249, 120, 110}

%%%%%%%%%%%%%%%%%%%%%%%%%%%%%%%%%%%%%%%%%%%%%%%%%%%%%%%%%%%%%%%%%%%%%%%%%%%%%%

%\if1\blind
\if0\blind
{
  \title{\bf   \textit{Counterfactual} Uncertainty Quantification of \textit{Factual} Estimand
of Efficacy from Before-and-After Treatment Repeated Measures Randomized Controlled Trials}
   \author{Xingya Wang\\ {\small https://orcid.org/0000-0003-1532-9952}, {\small xingya.wang-2@postgrad.manchester.ac.uk}\\
    {\small Department of Mathematics, The University of Manchester}\\
    Yang Han*\\ {\small https://orcid.org/0000-0002-4143-7765}, 
    {\small yang.han@manchester.ac.uk}\\
    {\small Department of Mathematics, The University of Manchester}\\
     Yushi Liu\\
     {\small liu\_yushi@lilly.com}\\
    {\small Eli Lilly and Company,}\\
    Szu-Yu Tang\\
    {\small Szu-Yu.Tang@pfizer.com}\\
    {\small Pfizer Worldwide Research \& Development}\\
%    and \\
   Jason C. Hsu*\\ {\small https://orcid.org/0000-0002-1277-0403}, 
   {\small Hsu.1@osu.edu}\\
   {\small Department of Statistics, The Ohio State University}\\
   }

\begin{comment}
     %% Information for the first author.
\author[]{Yujia Sun\inst{1}}
\address[\inst{1}]{Department of Mathematics, The University of Manchester}
%%%%    Information for the second author
\author[]{Yang Han\footnote{Co-corresponding author: {\sf{e-mail: yang.han@manchester.ac.uk}}, Phone: 44-161-2755806, Address: Department of Mathematics, University of Manchester, Oxford Road, Manchester, M13 9PL, UK.}\inst{1}}
%%%%    Information for the third author
\author[]{Haiyan Xu\inst{2}}
\address[\inst{2}]{Janssen Research and Development, Johnson \& Johnson}
%%%%    Information for the forth author
\author[]{Szu-Yu Tang\inst{3}}
\address[\inst{3}]{Pfizer Worldwide Research \& Development}
%%%%    Information for the fifth author
\author[]{Yushi Liu\inst{4}}
\address[\inst{4}]{Eli Lilly and Company}
\author[Sun {\it{et al.}}]{Jason C. Hsu\footnote{Co-corresponding author: {\sf{e-mail: jch@stat.osu.edu}}, Phone: 1-614-292-2866, Address: 1958 Neil Ave., Columbus, OH 43210, USA.}\inst{5}}
\address[\inst{5}]{Department of Statistics, The Ohio State University}
%%%%    \dedicatory{This is a dedicatory.}
\Receiveddate{zzz} \Reviseddate{zzz} \Accepteddate{zzz}
\end{comment}
 
  \maketitle
} \fi

\if1\blind
{
  \bigskip
  \bigskip
  \bigskip
  \begin{center}
    {\LARGE\bf \textit{Counterfactual} Uncertainty Quantification of \textit{Factual} Estimand
of Efficacy from Before-and-After Treatment Repeated Measures Randomized Controlled Trials}
\end{center}
  \medskip
} \fi

\bigskip
\begin{abstract}
This article quantifies the uncertainty reduction achievable for \textit{counterfactual} estimand, and cautions against potential bias when the estimand uses Digital Twins.  

Posed by \cite{neyman1923edited} who showed unbiased \textit{point estimation} from designed \textit{factual} experiments is possible, \textit{counterfactual} uncertainty quantification (CUQ) remained an open challenge for about one hundred years.  
The  $Rx: C$ \textit{counterfactual} efficacy we focus on is the ideal estimand for comparing treatment $Rx$ with control $C$, the expected outcome differential if each patient received \textit{both} $Rx$ and $C$.

Enabled by our new statistical modeling principle called ETZ, we show CUQ is achievable in Randomized Controlled Trials (RCTs) with \textit{Before-and-After} Repeated Measures, common in many therapeutic areas.
The CUQ we are able to achieve typically has lower variability than factual UQ.  

We caution against using predictors with measurement error, which violates regression assumptions and can cause \textit{attenuation} bias in estimating treatment effects.  
For traditional medicine and population-averaged targeted therapy, counterfactual point estimation remains unbiased.  
However, in both Real Human and Digital Twin approaches, estimating effects in \emph{subgroups} may suffer attenuation bias.

% % Original abstract from main paper
% The ideal estimand for comparing treatment $Rx$ with a control $C$ is the \textit{counterfactual} efficacy $Rx: C$, the expected differential outcome between $Rx$ and $C$ if each patient were given \textit{both}.  
% One hundred years ago, \cite{neyman1923edited} proved unbiased \textit{point estimation} of counterfactual efficacy from designed \textit{factual} experiments is achievable.  
% But he left the determination of how much might the counterfactual variance of this estimate be smaller than the factual variance as an open challenge.  
% This article shows \textit{counterfactual} uncertainty quantification (CUQ), quantifying uncertainty for factual point estimates but in a counterfactual setting, is achievable for Randomized Controlled Trials (RCTs) with \textit{Before-and-After} treatment Repeated Measures which are common in many therapeutic areas.  

% We achieve CUQ whose variability is typically smaller than factual UQ by creating a new statistical modeling principle called ETZ.  

% We urge caution in using predictors with measurement error which violates standard regression assumption and can cause \textit{attenuation} in estimating treatment effects. 
% Fortunately, we prove that, for traditional medicine in general, and for targeted therapy with efficacy defined as averaged over the population, counterfactual point estimation is unbiased.  
% However, for both Real Human and Digital Twins approaches, predicting treatment effect in \emph{subgroups} may have attenuation bias.  
\end{abstract}

\noindent%
{\it Keywords:}  Variability decomposition, Randomized Controlled Trials, Digital Twins, Counterfactual Uncertainty Quantification, ETZ modelling principle, Before-and-After Repeated Measures.%3 to 6 keywords, that do not appear in the title

\vfill

\newpage
\spacingset{1.45} % DON'T change the spacing!

\section{Motivation and scope of application}
When comparing a new treatment (which we denote by $Rx$) with a control (denoted by $C$), the ideal estimand is the \textit{counterfactual} efficacy, the expected differential outcome between $Rx$ and $C$ if each patient were given \textit{both} $Rx$ and $C$.  
Following \cite{rubin1978bayesian}, causal treatment effect is defined as
\begin{quote}
	``how the outcome of treatment compares to what would have happened to the same subjects under alternative treatment''
\end{quote}
by the 2019 International Council for Harmonization guideline ICHE9(R1) on Estimands\nocite{ICHE9(R1)}. 
From a patient's point of view, indeed what matters is the \textit{personal} treatment effect, the differential in outcomes if they switch from a control therapy $ C $ to a new therapy $ Rx $. 
We denote this as $Rx: C$ efficacy. 

Most \textit{factual} Randomized Controlled Trials (RCTs) are such that each patient is given \textit{either} $ Rx $ or $ C $ but not both.  
However, counterfactual \emph{point estimation} of counterfactual $Rx: C$ efficacy from factual studies is readily available, as random assignment of treatments to patients lets one prove that the point estimate of the differential $Rx$ and $C$ effects from modeling (\textit{average} treatment effect (ATE)) a factual RCT is unbiased for the \textit{counterfactual} $Rx: C$ efficacy.
%However, there is not a guidance for achieving estimating the factual estimand in counterfactual setting and there is potential risk in causing bias.

The proof that designed experiments allow for \emph{unbiased} \textit{counterfactual} point estimation of the \textit{average} treatment effect (ATE) from \textit{observable} outcomes can be traced to \cite{Neyman(1923)}, which has been edited and translated as \cite{Neyman(1990)}. 
%This long run patient response differential between $Rx$ and $C$ is the parameter of interest. 
% Importantly, Neyman realized that the variance of a \textit{factual} point estimate in standard practice is larger than what the variance of a \textit{counterfactual} point estimate would be, and considered the problem of accurately assessing the latter ``worthy of future study''.  
% See page 473 of \cite{Rubin(1990)} which commented on \cite{neyman1923edited}. 

Importantly, how to quantify the uncertainty in estimating a \textit{counterfactual} point estimate is a problem raised one hundred years ago but has remained unsolved.
Neyman considered this problem ``worthy of future study''.
Neyman is unable to assess the variance of a \textit{counterfactual} point estimate precisely, although he realized that the variance of a \textit{factual} point estimate in standard practice is larger than what the variance of a \textit{counterfactual} point estimate would be.
See page 473 of \cite{Rubin(1990)} which commented on \cite{neyman1923edited}. 

Current uncertainty quantification of this point estimate is based on the variability of differential $Rx$ and $C$ effects between \textit{different} patients, one treated with $Rx$ and the other with $C$, which intuitively is larger than the variability of differential effects if each patient were treated with both $Rx$ and $C$.  
With multiple components of variability, it is challenging to isolate and remove the person-to-person variability.  
Previous thinking was counterfactual \textit{uncertainty quantification} requires studies with \textit{crossover} designs and such which are rarely feasible.  
This article shows, actually, Counterfactual Uncertainty Quantification (which reduces the variance from factual uncertainty quantification) is achievable for a widely used kind of clinical trials that do \textit{not} involve \textit{crossover} designs.  

Disease areas listed in Table \ref{table:NofPatients} routinely conduct factual RCTs with each patient having \textit{Before-and-After} treatment Repeated Measures (without crossing over).
We call such studies factual BAtRM RCTs.  
We show, surprisingly, the uncertainty of $Rx: C$ efficacy point estimation from factual BAtRM RCTs can be quantified counterfactually.  
We do this by creating a new modeling principle, ETZ, to track individual patient outcomes in BAtRM RCTs and quantify their variability components.  
% BAtRM RCTs are standard practice in drug development for diabetes and neurological diseases. 
% We call these \textit{-after treatment} repeated measures studies (BAtRM studies for short.  
% impacts many therapeutic areas, although not all RCTs satisfies the requirement of BAtRM.
% How treatment outcome is measured depends on the disease.  
% In an oncology RCT, each patient has one measured Overall Survival (OS) time, for example.  
% On the other hand, in an Alzheimer disease RCT, functional measure such as Activity of Daily Living (ADL) is taken on each patient before randomization, and repeatedly after the treatment is given.  
% We call these latter RCTs \textit{-after treatment} repeated measures studies. 
Disease areas in which factual BAtRM RCTs are routine affect half a billion patients worldwide, as shown in Table \ref{table:NofPatients}.  

\begin{table}[hbtp]
	\centering
	\resizebox{\linewidth}{!}{
		\begin{tabular}{ccc}
			\toprule
			Therapeutic Area & Number of Patients (in millions) & Region \\
			\midrule \midrule
			Crohn’s disease and ulcerative colitis & 3 & U.S. \\
			\midrule
			Rheumatoid arthritis & 54 & U.S. \\
			\midrule
			Schizophrenia & 20 & worldwide \\
			\midrule
			Alzheimer’s disease & 50 & worldwide \\
			\midrule
			Type 2 diabetes & 400 & worldwide \\
			\bottomrule
	\end{tabular}}
	\caption{Therapeutic areas with \textit{Before-and-After} treatment Repeated Measures random controlled trials benefiting from the ETZ modeling principle. }\label{table:NofPatients}
\end{table}

Efficacy is primarily assessed at what is called the \textit{milestone} visit. % xw 20240331 define milestone visit
One can assess efficacy based on either the outcome measured at the milestone visit or \textit{change-from-baseline} defined as the difference between the outcomes at the milestone visit and at baseline.
The current practice in BAtRM RCTs uses \textit{change-from-baseline}.  

To gain insight, 
we propose a new statistical modeling principle called ETZ which separates variability of patient outcomes at the milestone visit into three components: variability in severity (stage of the disease that patients are initially at) $Z$,
%before treatments
variability in trajectories $Traj$ of patients' responses to treatments, and variability in the measurement errors $E$. 
% ETZ is non-standard in that $Z$ is not directly observable.  

Our ETZ modeling principle in Section \ref{sec:CUQ} shows using \textit{change-from-baseline} instead of the milestone visit outcome reduces the uncertainty in efficacy assessment provided that the measurement error variability is not the main source of variability at the baseline visit (visit 1), a minimal requirement (for otherwise the outcome scale is of questionable usefulness).  
So the ETZ modeling principle validates current practice.  
%If Var(E)>Var(Z), the scale used is not reliable. For first visit, baseline = Z + E. E dominates baseline when Var (E) > Var (Z).
% If one assesses $Rx: C$ efficacy using only data at the milestone efficacy assessment time point, then variability of the estimate contain (2$\times$) patient-to-patient variability and (2$\times$) within patient measurement variability. 
% Current practice for BAtRM RCTs is to estimate $Rx: C$ efficacy using \textit{change-from-baseline} measurements which has no patient-to-patient variability but (4$\times$) within patient measurement variability.
%Q: Shall we only consider continuous outcome here?
% So whether this is good practice depends on whether patient-to-patient variability is bigger or smaller than within patient measurement variability, which our newly proposed ETZ modeling can determine.  

Further, without changing the (unbiased) treatment efficacy point estimation in current practice, our ETZ modeling principle allows what we call Counterfactual Uncertainty Quantification (CUQ) which calculates a (typically smaller) quantified variability of an idealized counterfactual estimand instead of the factual estimand. 
%With the standard practice of measuring treatment outcome as change-from-baseline effectively removing the before-treatment variability $Z$, we show that the larger the variability of measurement error $E$ relative to the variability of trajectory $T$ the larger the size of this reduction. 
We illustrate CUQ using realistic examples of BAtRM RCTs. 

A final insight from the ETZ modeling principle is that, since $Z$ is unobservable, it must be estimated. 
While \textit{baseline} is often used, $Z$ can more generally be estimated from a set of biomarkers.  
However, when predictors are measured with error, standard regression analysis leads to \textit{attenuation} of treatment effect estimates.
Fortunately, we found that \textit{counterfactual} point estimation remains unbiased for traditional medicine. For targeted therapies, it is also unbiased, as long as efficacy is defined as population-averaged (see Section~\ref{sec:CFEUnbiased}). 
That said, all methods including our self-controlled human design and digital twins—should note that subgroup-level effect estimation may be biased if biomarkers are measured with error (see Section~\ref{sec:CFbiased}).

\section{ETZ, a modeling principle}\label{sec:ETZ}

% \subsection{Causal Estimand definition of treatment effect}
%Note: I suggest that name of this section needed to be more general

In practice, the differential effect between $ Rx $ and $ C $ is typically estimated from outcomes observed with some patients given $ Rx $ and others given $ C $. 

% The proof that designed experiments allow for \emph{unbiased} \textit{counterfactual} point estimation of the \textit{average} treatment effect (ATE) from \textit{observable} outcomes can be traced to \cite{Neyman(1923)}, which has been edited and translated as \cite{Neyman(1990)}. 
% %This long run patient response differential between $Rx$ and $C$ is the parameter of interest. 
% Importantly, Neyman realized that the variance of a \textit{factual} point estimate in standard practice is larger than what the variance of a \textit{counterfactual} point estimate would be, and considered the problem of accurately assessing the latter ``worthy of future study''.  
% See page 473 of \cite{Rubin(1990)} which commented on \cite{neyman1923edited}.  
% We are happy to report that, in the broadly applicable setting of BAtRM RCTs, the ETZ modeling principle enables this assessment.
While it is known that randomization allows unbiased estimation of a Counterfactual estimand from factual observations, Counterfactual Uncertainty Quantification (CUQ) of such an estimate has been largely thought as requiring crossover or N-of-1 studies since \cite{neyman1923edited}.  
We are happy to report that, in the broadly applicable setting of BAtRM RCTs, the ETZ modeling principle enables CUQ.  

\subsection{Conceptualizing an ideal \textit{counterfactual} efficacy}\label{sec:ignorability.proof}

Let $ Trt_i $ be the random variable that assigns the $ i^{th} $ patient to either $ Rx $ or $ C $.
%add the superscript for visit
For the $ i^{th} $ patient, let $ Y_i (Rx) $ and $ Y_i (C) $ represent \emph{potential} outcomes if that patient is assigned to $ Rx $ or $ C $ respectively.
Let $ X $ represent relevant covariates in general.
%Our discussion focuses on the simplest but important special case that $ X = g^- $ or $ g^+ $.
The \textit{mean} effect of a treatment $ Trt $ is the expected potential (possibly \textit{counterfactual}) outcome if \emph{all} individuals in the population had received the treatment $ Trt $.

What is called an ``ignorable'' random assignment of patients to $ Rx $ and $ C $ is one such that, for each given $ X = x $, the random assignment variable $ Trt_i $ is \emph{independent} of the \textit{potential} outcomes $ Y_i (Rx) $ and $ Y_i (C) $.

%Let $Y^T_{r}$ represent the observed outcome of the $ r^{th} $ patient given the treatment $ Trt $, $ T = Rx $ or $ C $.

%For the $ i^{th} $ patient, let
%$ Y_i(Rx) $ and $ Y_i(C) $ denote the \emph{potential} outcomes given $ Rx $ and $ C $.

For the $ r^{th} $ patient assigned to $ Rx $, let $Y^{Rx}_{r}$ represent the \emph{observed} outcome,
and for the $ s^{th} $ patient assigned to $ C $, let $Y^{C}_{s}$ represent the \emph{observed} outcome.

Only $ Y_r^{Rx} $ and $ Y_s^{C} $ are observable, not $ Y_r(C) $ or $ Y_s(Rx) $.
However, since $ Trt_i $ is independent of potential outcomes under ignorable randomization,
conditional on $ Trt_i $ being either $ Trt_i = Rx $ or $ Trt_i = C $, potential outcomes and observable outcomes have the same expectations:
\begin{eqnarray*}
	& & \mathbbm{E} [Y_i(Rx) | X_i = x_i ] \\
	& = & \mathbbm{E}\{ \mathbbm{E} [ Y_i(Trt_i) | Trt_i = Rx, X_i = x_i ] \}\\
	& = & \mathbbm{E} [ Y^{Rx}_i | X_i = x_i ],\\
	%\end{eqnarray*}
	%\begin{eqnarray*}
	& & \mathbbm{E} [Y_i(C) | X_i = x_i ] \\
	& = & \mathbbm{E}\{ \mathbbm{E} [ Y_i(Trt_i) | Trt_i = C, X_i = x_i ] \}\\
	& = & \mathbbm{E} [ Y^{C}_i | X_i = x_i ].
\end{eqnarray*}
%\begin{eqnarray*}
%            & & E [Y_r(Rx) | X_r = x_r ] \\
%            & = & E\{ E [ Y_r(Trt_r) | Trt_r = Rx, X_r = x_r ] \}\\
%            & = & E [ Y^{Rx}_r | X_r = x_r ],\\
%%\end{eqnarray*}
%%\begin{eqnarray*}
%            & & E [Y_s(C) | X_s = x_s ] \\
%            & = & E\{ E [ Y_s(Trt_s) | Trt_s = C, X_s = x_s ] \}\\
%            & = & E [ Y^{C}_s | X_s = x_s ].
%\end{eqnarray*}
%Therefore,
%$$ Pr(Y(m) | X) = Pr(Y^{m} | X; T = m) $$
%\begin{eqnarray*}
%Pr(Y_i(Rx)|X)  & = & Pr(Y_i^{Rx} |X) \\
%Pr(Y_i(C) |X) & = & Pr(Y_i^{C} |X)
%\end{eqnarray*}

% \subsection{Defining efficacy}

Assuming the distribution of $ X $ is the same for $ Rx $ and $ C $, taking expectation over the distribution of $X$, 
if efficacy $ \tau $ is defined as the expected \emph{difference} of \emph{potential} outcomes, then
\begin{eqnarray}
	\tau & = & \mathbbm{E} [ Y(Rx) - Y(C) ] \nonumber \\
	& = & \mathbbm{E}_X [ \mathbbm{E} ( Y(Rx) - Y(C)  | X = x ) ] \nonumber \\
	& = & \mathbbm{E}_X [\mathbbm{E} ( Y(Rx) | X = x ) - \mathbbm{E}( Y(C)  | X = x ) ] \label{eq:difference_potential_efficacy}\\
	& = & \mathbbm{E}_X [ \mathbbm{E} ( Y^{Rx} | X = x ) - \mathbbm{E} ( Y^{C} | X = x ) ] \label{eq:efficacy_subgroup} \\
	& = & \mathbbm{E}_X [ \mathbbm{E} ( Y^{Rx} | X = x ) ] - \mathbbm{E}_X [ \mathbbm{E} ( Y^{C} | X = x ) ] \label{eq:propensity_equal} \\
	%& = & E_X [ E ( Y[Rx] | Trt_i = Rx, X = x ) - E ( Y^{C} | Trt_i = C, X = x ) ] \label{eq:efficacy_subgroup} \\
	%& = & E_X [ E ( Y[Rx] | Trt_i = Rx, X = x ) ] - E_X [ E ( Y^{C} | Trt_i = C, X = x ) ] \label{eq:propensity_equal} \\
	& = & \mathbbm{E} [Y^{Rx}] - \mathbbm{E} [Y^C] \label{eq:difference_outcome_efficacy}
\end{eqnarray}
so this efficacy in the population space with infinitely many patients can be assessed based on \textit{observable} outcomes. 
%However, before averaging unconditionally over the distribution of $X$, there may be specific $X$ values for which one should consider conditional efficacy.  

\section{\texorpdfstring{Conceptualizing an ideal personal $Z$ covariate}{Conceptualizing an ideal personal Z covariate}}
The ETZ modeling principle conceptualizes each patient as having their own random \textit{intercept} $Z$ before any treatment is given.  
\textit{Trajectory}, denoted by $ Traj $, is the random response from each patient to the treatment $ Trt $ given.
Within each $ Trt $, trajectory $ Traj $ is randomly distributed 
around a \textit{profile} of long run average trajectories (without a particular shape specified).  
% linear, quadratic, Emax, or whatever).  
ETZ is a \textit{counterfactual} modeling principle, with observable outcomes given the factual treatment, and potential outcomes given the \textit{counterfactual} treatment.  

Compared to the usual \textit{baseline} predictor, the $ Z $ component has two unique features: it is unobservable and without measurement error.
As visualized in Figure \ref{fig:ETZ3RCT}, neither the intercept $Z$ nor the 
trajectory curves are directly observable, because there are measurement errors.  
% Observed, or potentially observed, are $ Y^{[v]}_{i}, ~v = 1, \ldots, m $, represented as dots and squares in Figure \ref{fig:ETZ3RCT}.
Observed and potential outcomes are represented as dots and squares respectively in Figure \ref{fig:ETZ3RCT}.

We call ETZ a \textit{counterfactual} modeling principle because $ Z $ is not observable.  
Each individual is assigned a scalar $ Z $ for their covariate $ X $, on the $ Y $ outcome scale.
The ETZ modeling principle uses $Z$ to model biological variability among patients, conceptualizing initial intercept value $Z$ as predictive of outcome.  

The ETZ modeling principle reduces the quantified uncertainty of estimated \textit{counterfactual} efficacy at the population level by first using each patient as their own control % canceling the effect of $Z$. 
(as in current practice), then further reduces it by calculating counterfactually under the ETZ modeling principle.  
% We believe the resulting sample size reduction we achieve is at least as competitive as any other approaches, perhaps more so.  

The ETZ modeling principle cannot be applied directly since $ Z $ is unobservable.  
Model fitting has to be done factually, so (for proof-of-concept) we model the data as Mixed Model Repeated Measures (MMRM) 
% Since $Z$ is not observable, 
using \textit{baseline} measurement as a surrogate to $ Z $.  
We then make a 1-to-1 transformation between the variability components in MMRM and ETZ.  
We call this process Real Human Counterfactual Estimation (RHCE).  

\subsection{A factual Digital Twins model}
Instead of conceptualizing patients as having different (unobservable) initial values $ Z $, one can also account for patient variability by their having different observable biomarker values $ (b_1, \ldots, b_p) $ such as age, baseline or other measurement of disease severity, sex, etc.  
An algorithm $ g $ can be trained to predict an $ Rx $ patient's digital twin's $ C $ outcome and a $ C $ patient's digital twin's $ Rx $ outcome based on $ (b_1, \ldots, b_p) $.  
In what we call Digital Twin Counterfactual Estimation (DTCE), after calculating the difference between each patient's factual and its digital twin's \textit{counterfactual} outcome, 
population $Rx: C$ efficacy is estimated by averaging them across the patients.  
See \cite{NAS(2024)DT}, for example, for such thinking.  
DTCE can also potentially reduce uncertainty of estimated efficacy. 

\paragraph*{\textbf{Beware of covariates with measurement error}}

It is important to be cognizant of covariates measured with error (e.g, baseline or other measures of disease severity).  
Predicting outcomes based on such covariates violates the standard regression modeling assumption that predictors are measured without error.  

Inference from models with measurement error in predictors, called errors-in-variables models, % (the $Z$ in ETZ), 
will have \emph{attenuation} bias.  
Therefore, in Section \ref{sec:CFEUnbiased}, we explain the care that both RHCE and DTCE need to take to avoid this bias.  
% To place both RHCE and DTCE in the same framework, we take the algorithm $ g $ as being trained to create a digital patient value $ B = g(b_1, \ldots, b_p) $ similar to \textit{baseline}, to predictor that patient's $ Y $ outcome with, given either $ Rx $ or $ C $.  
To place both RHCE and DTCE in the same framework, we take the algorithm $ g $ as being trained to create a digital patient value $ B = g(b_1, \ldots, b_p) $ similar to \textit{baseline}, which is used as one predictor for patient's $ Y $ outcome, given either $ Rx $ or $ C $. 

\subsection{ETZ notations}
Index each of the $ m $ visits of a repeated measures study by $ v $, $ v = 1, \ldots, m$.  
Denote by $ Y^{Trt[v]} $ the outcome of a patient given treatment $ Trt $ (either $ Rx \mbox{ or } C $) measured at visit $ v $.  
Values of profiles for $ Rx $ and $ C $ at times of visits are the fixed effects in the MMRM model (\ref{model:MMRM}). 
%Similar to the ETZ model, MMRM does not assume a functional form for the profiles.  
The ETZ modeling principle does not assume a functional form for the profiles, neither does MMRM.

\begin{figure}[hbtp]
	\centering
	\includegraphics[width=0.7\linewidth]{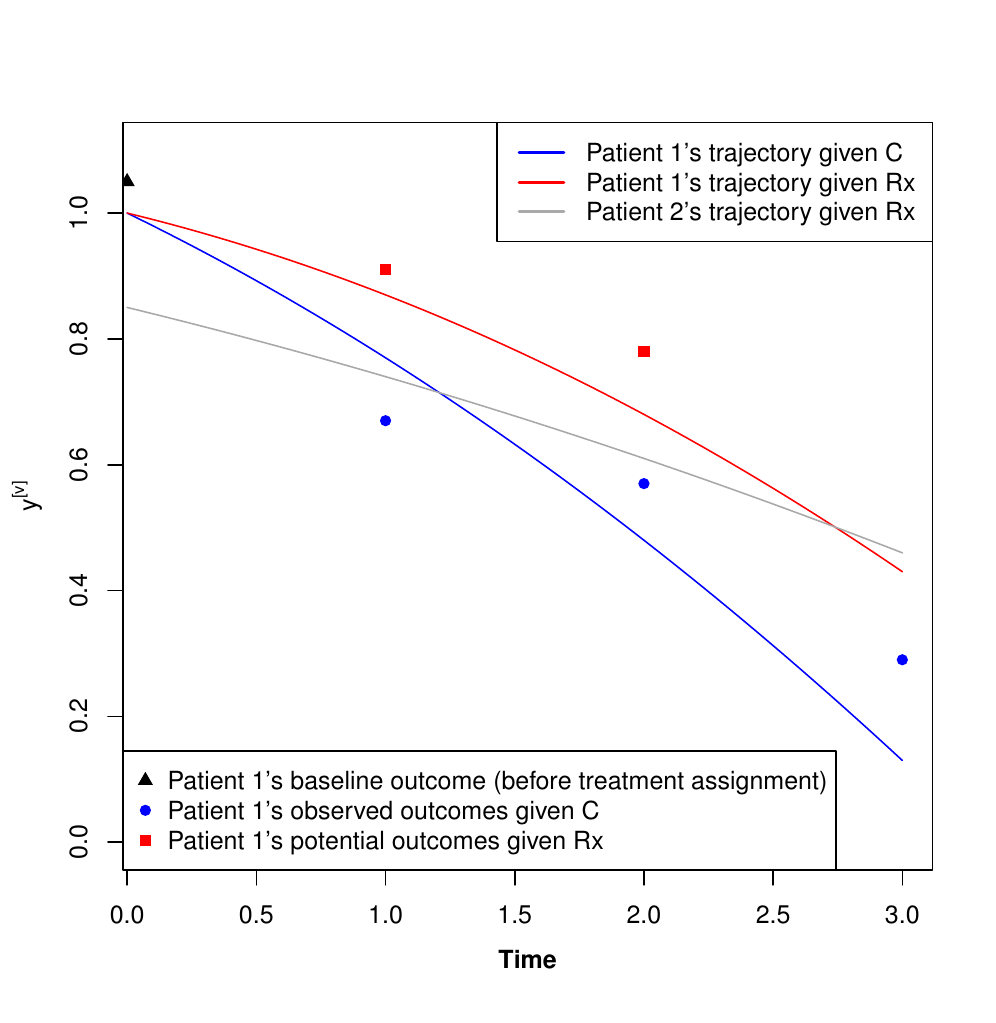}
	\caption{Components of the ETZ modeling principle for a RCT with $ m = 4$ visits at Time = 0, 1, 2, and 3. Black triangle is the baseline of patient 1 before any treatment is taken. Blue curve is the trajectory of patient 1 who is assigned to $ C $, and the blue dots are the observed $ Y $.  Red curve represents patient 1's potential trajectory had they been assigned to $ Rx $, and the red squares are the potential $ Y $ (with last value missing).  Grey curve is the trajectory of patient 2 who is assigned to $ Rx $.  Measurement errors are differences between the outcomes and their trajectories.}
	\label{fig:ETZ3RCT}
\end{figure}

To be more specific, conceptualize $ Y^{Trt[v]} $ as having either two or three random components.  
At visit 1 ($ v = 1 $), before any treatment is given, $ Y^{Trt[1]} $ has two random components, a component $ Z $ which is that patient's intercept (true baseline measurement), and a measurement error component $ E $ which is the difference between the true and the observed baseline.  
Thereafter, $ Y^{Trt[v]} $ has three random components, a $ Z $ which is that patient's intercept (true baseline measurement), plus a random trajectory $ Traj $ component (around its mean profile $ \mu $) representing that patient's response to the treatment given up to the time of visit $ v $ ($ v > 1 $), and a measurement error component $ E $ which is the difference between the measured outcome and that patient's true trajectory at each visit $ v $.  
These components can be seen in Figure \ref{fig:ETZ3RCT}.  
%If we make the typical ANOVA and multiple comparisons assumption that, at a particular visit (in a RCT), variability is sufficiently constant across treatments $ i $, then 
% Distinct from MMRM (\ref{model:MMRM}), the ETZ model clinically separates measurements $ Y^{[1]}_{i} $ at visit 1 ($ v = 1 $) before any treatment is given from measurements $ Y^{Trt[v]} $ at visits $ v > 1 $ once treatments are given.  
% See (\ref{eq:Ybefore}) and (\ref{eq:Yafter}) below.  xw 20240327 move the difference between ETZ and MMRM to later part
It is assumed that measurement error $ E $ is independent of $ Z $ and $ Traj $, and i.i.d. over the visits, across all the patients.  
%We center $ E^{[1]}, E^{[m]} $, and $ Z^{[v]} $ to have expectation zero, 
%and assume that they are independent.  
% , with distributions that do not depend on treatment $ i $.  

\subsection{\textit{Independence} as a starting point}\label{sec:IndCase}

Our discussion of the ETZ modeling principle starts in the setting where the random intercept $ Z $ and the random trajectory $ Traj $ are \textit{independent} within each treatment arm and goes toward where the direction of dependence between $ Z $ and $ Traj $ is known (with consequence to be discussed in Section \ref{sec:ETZDecomposition}).
% We assume the random intercept and the random trajectory (within each arm) are \textit{independent}, 
% interaction term between random intercept and random slope is not testable, 1 observed profile per patient  
% with intercept being baseline, trajectories over time would be parallel, sicker patients end up still being sicker 
% this is the rationale for baseline covariate adjustment with no interaction term with treatment 
% biomarker for targeted therapy might predict trajectory with sicker patients getting more benefit 
% biomarker covariate adjustment should have interaction term with treatment 

\textit{Independence} means that, the true stage of the disease that a patient is at before they receives either $ Rx $ or $ C $ does not inform on whether that patient's response trajectory is going to be above or below the profile $ \mu $ (long run average of the trajectories). 
By the Monotone Class Theorem, it is sufficient to check whether $ P\{B|A\} = P\{B\}$ for events $ A $ and $ B $ of the form $ B = \{improve  > b\} $ and $ A = \{severity > a\}$.  
That is, within each treatment arm, whether knowledge of severity being greater than 25 (say) informs on whether a patient would improve by more than 10 points (say).
%\footnote{By the Monotone Class Theorem, it is sufficient to examine events of this form.}  
%The concept of baseline being prognostic will be defined in Section \ref{sec:GenesisBaseline}.  jch 29Apr2025
(Note that independence between the random effects $ Z $ and $Traj$ \textit{within} each treatment arm is unrelated to whether the fixed effect response profiles are parallel or not.)   

% In addition, while trajectories for $ Rx $ and $ C $ are expected to have different profiles, we assume \textit{variabilities} of the trajectories are the same for $ Rx $ and $ C $. 
% Assuming independence between a patient's random intercept and that patient's random response trajectory given a particular treatment means that, while severity may be \textit{prognostic} of outcome at the milestone visit, it is not \textit{predictive} of the response trajectory (whether a slope would be above or below the average of the slopes, if the trajectories are straight lines, for example).  
% From a biological perspective, the independence assumption means true severity should not be a drug targeting biomarker (that directly measures the amount of drug targets available for the compound to act upon, for example).  %Need further discussion xw 20240417

If patient-level data is available, then one way to check independence is to see how baseline measurement is predictive of \textit{change-from-baseline}.  
We take an AbbVie Alzheimer's disease (AD) data as an example.
\cite{schnell2017subgroup} described an Alzheimer Disease study from AbbVie which compared three doses of a new treatment (doses 1, 2, and 3) with a negative control (dose 0, a placebo).  
There was an active control (dose 4) in the study as well.  
Outcome measure was ADAS-cog11, and duration of this Phase 2 study was 24 weeks.
%For the AbbVie AD data (to be further analyzed in Section \ref{sec:EZonAbbVie}), estimated correlation between baseline (variable ``severity'' in the data set) and change-from-baseline (negative of the ``improve'' variable in the data set) is 0.16, with an R-square of 0.025 from linear regression.  
%For the AbbVie AD data, the maximum correlation between baseline (severity) and improvement ($ - $change) for the 3 doses was 0.17, and the maximu R-square regressing improvement on severity was 0.03.  
For the AbbVie AD data, the R-square regressing \textit{change-from-baseline} (negative of the ``improve'' variable in the data set) on baseline (variable ``severity'' in the data set) allowing different profiles for the placebo and the 4 doses is 0.17.  
So it appears severity is not very predictive of trajectory.

A subtle concern is, regression modeling assumes predictors are measured without error, what if some of them have measurement error?     
Fortunately, as will be explained in Section \ref{sec:CFEUnbiased}, as long as efficacy is defined as “on average”, estimated $Rx: C$ efficacy is unbiased.   
However, that Section \ref{sec:CFbiased} also shows, in predicting efficacy in subgroups or at particular biomarker values, 
ignoring measurement error in predictors % such as MMSE and $\beta$Amyloid load have 
will lead to bias. 

\subsection{Properties of a \textit{counterfactual} model}
If the outcome of $i^{th}$ patient given treatment $Trt_i$, $Y_i(Trt_i) $, is thought to only depend on the treatment $ Trt_i $ and the covariate $ X_i $, then the model is said to be \textit{deterministic} \textit{counterfactual}.

In real life, outcome measure on a patient is non-deterministic. 
Each potential measurement on each patient, if replicated, would be somewhat different. 
Reason for such variability includes the patient having a good day or a bad day (if they are asked to perform some tasks), rater variability, measurement error in analytic chemistry. 
If, in addition to the treatment effect $ Trt_i $ and the covariate effect $ X_i $,  there is random measurement variability affecting $ Y_i(Trt_i) $, then the model is said to be \textit{non-deterministic} \textit{counterfactual}. 
In standard statistical notations, we write such a model as
\begin{equation}
	Y_i(Trt_i) = h(Trt_i, X_i) + \epsilon_i \label{eq.CounterFact}
\end{equation}
where $ h(Trt_i, X_i) $ represents treatment and covariate effects and $ \epsilon_i $ represents measurement variability with $ E(\epsilon_i) = 0$, assuming $ \epsilon_i $ is independent of $ (Trt_i, X_i) $.  
Since $ E(\epsilon_i) = 0$, the derivation leading to the conclusion (\ref{eq:difference_outcome_efficacy}) that $ \tau = E (Y^{Rx}) - E (Y^{C}) $  remains valid. 

The \textit{counterfactual} model represents an ideal situation, with the following properties. 

\begin{description}
	\item[Unbiased efficacy] With each patient receiving both $Rx$ and $C$, the $X$ values are perfectly matched between the $Rx$ and $C$ outcomes.  So $ E[Y(Rx) - Y(C)] $ is not biased by patients having high $X$ values systematically tending to receive $Rx$ while patients with low $X$ values tending to receive $C$ (say). 
	\item[Reduced variability] Intuitively, $ Y(Rx) - Y(C) $ with both $Rx$ and $C$ given to the same patient has less variability than $ Y^{Rx} - Y^{C} $ giving $Rx$ to one patient and giving $C$ to a different patient. 
\end{description}
The ETZ modeling principle we propose in this article leads to model (\ref{eq.CounterFact}) that separates measurement variability $ \epsilon_i $ (the ``E'') from patient-to-patient variability in $ h(Trt_i, X_i) $ and crucially further separates the latter to before treatment (the ``Z'') and after treatment (the ``T'') variability. 

%\subsection{Baseline and milestone studies for \textit{counterfactual} inference}
\subsection{BAtRM for \textit{counterfactual} inference}
%Intuitively, if one were able estimate counterfactual within-patient differentials, then variability of the estimated differential treatment effect would be smaller than from a pair of different . 
To achieve \textit{counterfactual} inference by giving \textit{both} $Rx$ and $C$ to each patient is unethical for diseases that cause mortality (e.g., cancer) or irreversible mobility (e.g., arthritis).%\footnote{Pk/Pd studies with crossover designs for establishing bioequivalence involve healthy subjects, not patients.} % jch 24Jul2024 commenting out since StatSci does not allow footnotes

Consider instead a factual RCT in which each patient first gets a baseline measurement $ Y^{[1]} $ before being randomized to receive either $ Rx $ or $ C $, and then another measurement $ Y^{[m]} $ at the milestone time point when efficacy in the primary endpoint is assessed. 
\textcolor{black}{We call such RCTs \textit{Before-and-After} treatment Repeated Measures (BAtRM) studies.} 

Therapeutic areas with such RCTs include type 2 diabetes mellitus (T2DM), neurodegenerative diseases such as Alzheimer's disease (AD), psychiatric diseases such as schizophrenia, immunological diseases including rheumatoid arthritis, ulcerative colitis, Crohn's disease, and psoriasis, afflicting hundreds of millions of patients worldwide in total. 

\textit{Counterfactual} inference includes two aspects: \textit{counterfactual} (point) estimation and \textit{counterfactual} uncertainty quantification.
Surprisingly, \textit{counterfactual} inference is possible from factual baseline and milestone RCTs, following a 3-step process. 
The first step achieves \textit{counterfactual} estimation by Gauss-Markov theorem using factual sample, while the rest two steps explains the variance reduced by \textit{counterfactual} uncertainty quantification from factual case.
\begin{description}
	\item[Balancing] For a finite factual sample (which will not have perfectly balanced, let alone perfectly matched, $X$ values between $Rx$ and $C$), Gauss-Markov covariate adjustment ensures unbiased \textit{counterfactual} point estimation by mapping the sample space with imbalance to an ideal balanced \emph{population} space, as we demonstrate in Section \ref{sec:CFbiased}. 
	\item[Baselining] In RCTs with baseline and milestone measurements, variability incurred by giving $Rx$ to one set of patients and $C$ to the other patients can largely be removed by measuring outcome as \textit{change-from-baseline} ($ Y^{[m]} - Y^{[1]}$), as we show in Section \ref{sec:CUQ}. 
	\item[Self-controlling] % In deciding whether $Rx$ is better than $C$,
	Merely providing an unbiased point estimate of efficacy with reduced variability is insufficient for statistical inference, an accurate assessment of the variability of the estimate is needed as well. 
	Section \ref{sec:ETZDecomposition} shows how to transform the following variabilities from a factual model
	\begin{description}
		\item[$ \sigma^{B} $] Standard deviation of baseline measurements
		\item[$ \sigma^{Mile} $] Standard deviation of outcome at the milestone time point
		\item[$ \sigma^{Change} $] Standard deviation of \textit{change-from-baseline}
	\end{description}
	to the three variabilities in the ETZ modeling principle, from which variability of our \textit{counterfactual} estimand can be calculated. 
\end{description}

% Even though a factual study may have more visits than just for baseline and milestone measurements,
ETZ is a modeling\textit{ principle} applicable to any BAtRM data structure, not restricted to any specific analysis method.  
How to estimate $ \sigma^{B} $,  $ \sigma^{Mile} $, and $ \sigma^{Change} $ from factual data is not unique. 
For proof of concept, we show how they can be obtained from widely used 
Mixed Model Repeated Measures (MMRM) analyses introduced in Section \ref{sec:MMRM}, and then transformed 1-to-1 to variabilities in the ETZ modeling principle in Section \ref{sec:ETZDecomposition}.

% \section{Factual modeling of BA\MakeLowercase{t}RM studies}
\section{Factual modeling and \textit{counterfactual} uncertainty quantification}

The setting considered by MMRM is a longitudinal Randomized Controlled Trial in which there are $m$ planned visits (visit 1, $ \ldots, m$) for all the patients. 
% The setting of this article is a longitudinal Randomized Controlled Trial in which there are $m$ planned visits (visit 1, $ \ldots, m$) for all the patients. 
The times of these visits after randomization are the same for all the patients, and visit $m$ is the milestone time point at which efficacy in the primary endpoint is assessed. 
% Index each of the $ m $ visits of such a study by $ v $, $ v = 1, \ldots, m$. 
Denote by $ Y^{[v]}_{i} $ the measurement on patient $ i $ during visit $ v $. 

\subsection{Mixed Model Repeated Measures (MMRM)}\label{sec:MMRM}
The so-called Mixed Model Repeated Measures (MMRM) representation is % SAS Mixed book page 271
\begin{align}\label{model:MMRM}
	Y_{i}^{[v]} = \mu + \alpha^{[v]} + \beta^{Trt} + (\alpha\beta)^{Trt[v]} + b_{i}^{Trt} + w_{i}^{Trt[v]},&\\
	~ v = 1, \ldots, m.&\nonumber
\end{align}
Fixed effects are represented by $ \mu + \alpha^{[v]} + \beta^{Trt} + (\alpha\beta)^{Trt[v]} $, the mean response under treatment $ Trt $ at time $ v $, where $ \alpha^{[v]} $ is the visit (time) effect, $ \beta^{Trt} $ is the treatment effect, and $ (\alpha\beta)^{Trt[v]} $ is the treatment$ \times $visit interaction. 
Random between-subject effect for the $ i^{th} $ patient assigned to treatment $ Trt $ is $ b_{i}^{Trt} $, while random within-subject effect is represented by $ w_{i}^{Trt[v]} $. 

In an MMRM analysis, corresponding to the REPEATED statement in Proc Mixed of SAS, \textit{visit} (time) is not considered a continuous variable but rather just a label, a \texttt{CLASS} variable in SAS or a \texttt{factor} in R. 
No profile of the response over time is modeled, and a \texttt{Type =} option lets the user specify a structure for the $ m \times m $ variance-covariance matrix of measurements on the patients across the $ m $ visits. 
Given patient level data, Proc Mixed provides an estimated variance-covariance matrix $\mbox{\boldmath $R$}_{m \times m}$ for it, from which
% $ \hat{\sigma}^{[1]} $ and $ \hat{\sigma}^{[m]}$,
\begin{align}
	\sigma^{B} & =  \sqrt{Var(Y_{i}^{[1]})} \nonumber \\
	\sigma^{Mile} & =  \sqrt{Var(Y_{i}^{[m]})} \label{eq.3numbers}\\
	\sigma^{Change} & =  
	\sqrt{Var(Y_{i}^{[m]}) + Var(Y_{i}^{[1]}) - 2Cov(Y_{i}^{[m]}, Y_{i}^{[1]})} \nonumber
\end{align}
can all be estimated. 
By default, Proc Mixed displays the covariance matrix $\mbox{\boldmath $R$}_{m \times m}$ for the first subject's visits that are not missing.  
In case the first subject is missing Visit $m$ value, 
one can get the covariance matrix of a subject with Visit $m$ value using the \texttt{R=} option in the \texttt{REPEATED} statement.  

%With $\mathbf {Y}_i$ denoting for patient $i$ values of outcome from the first visit (baseline value) to the milestone visit, a statistical model for repeated measures data is
%\begin{eqnarray}\label{Mean}
%            \mathbf{Y}_i=\left[\begin{array}{c}y_{i 1} \\ y_{i 2} \\ y_{i 3} \\ y_{i 4} \\ \vdots \\ y_{i n_i} \end{array}\right], \operatorname{E}(\mathbf{Y}_i)=\mathbf{X}_i \mathbf{\bbeta},
%\end{eqnarray}
%with $\mathbf{Y}_i$'s for different patients being independent each having variance-covariance structure
%\begin{eqnarray}\label{model} \operatorname{Var}[\mathbf {Y}_i]=\left[\begin{array}{cccccccc}\sigma_{1}^{2} & \sigma_{12} & \sigma_{13} & \sigma_{14} & \cdots & \sigma_{1n_i} \\ &\sigma_{2}^{2} & \sigma_{23} & \sigma_{24} & \cdots & \sigma_{2n_i} \\  &&\sigma_{3}^{2} & \sigma_{34} & \cdots & \sigma_{3n_i}\\ & & & \sigma_{4}^{2} &\cdots& \sigma_{4n_i}\\ &  &  &  & \ddots &\vdots\\ & &&&&\sigma_{n_i}^{2} \end{array}\right]. 
%\end{eqnarray}
%The dimension of $\mathbf{\Sigma}_i$ is the total observation times for each patient and $n_i$'s can be different because of the missing data issue.
% \subsection{Idea}
% \subsection{Connection between MMRM and ETZ model}
The standard MMRM analysis from which we obtain these estimates assumes the missing data mechanism is MAR (Missing At Random).  
Strategies for dealing with other missing mechanisms are given in Chapter 19 of \cite{Mallinckrodt&Lipkovich(2019)}.  

\subsection{Comparison between MMRM and ETZ modeling}
The ETZ modeling principle conceptualizes BAtRM clinical data biologically, with each of $ E $, $ Traj $, and $ Z $ medically interpretable.  
ETZ separates the between subject effect $ b_i^{Trt} $ in the MMRM model (\ref{model:MMRM}) into two parts: a visit 1 before-treatment $ Z $ part, and a treatment-dependent $ Traj $ part starting with visit 2.  
ETZ lets us articulate control of variability.  
Specifically, while intercept variability $ Var(Z) $ can be reduced by narrowing the patient entry criterion, trajectory variability $ Var(Traj) $ can be reduced by decreasing drug target heterogeneity among patients.  
Variability in measurement error $ Var(E) $ can be reduced by using better trained raters, for example. 

Mixed Model Repeated Measures (MMRM) is not necessary for the ETZ modeling principle but only one of possible methods which provides ETZ modeling with the information needed.
We use MMRM for proof-of-concept as it is widely used in drug development currently.
However, other methods such as fitting a random coefficient model (RCM) when appropriate can be used alternatively.

%ETZ modeling principle is a brand-new modeling principle, although there are some connections between ETZ and MMRM.
%$Baseline = Z + E$ is considered as fixed effect in MMRM while $ Z $ is considered as random component in ETZ modeling principle.
%The $ Traj $ (T) and $ Z $ parts of ETZ separate the between subject effect $ b_i^{Trt} $ in the MMRM model (\ref{model:MMRM}) into two parts: a visit 1 pre-treatment %-na\"{i}ve $ Z $ 
%part, and a treatment-dependent $ Traj $ part starting with visit 2.  
%The measurement error $ E $ part of ETZ is different from the within subject measurement error in model (\ref{model:MMRM}).  
% Each of $ E $, $ Traj $, and $ Z $ is medically interpretable, and their variabilities can be controlled to varying extent.  
% For example, while intercept variability $ Var(Z) $ can be reduced by narrowing the patient entry criterion, trajectory variability $ Var(Traj) $ can be reduced by decreasing drug target heterogeneity among patients.  
% Variability in measurement error $ Var(E) $ can be reduced by using more highly trained raters, for example.  

% High variability of \textit{change-of-baseline} separation between $ Rx $ and $ C $ causes low rate of successful confirmatory studies.  
% How much of an impact on $ Var(Y^{[\mathrm{change}]}_{i}) $ will reduce each of $ Var(Z) $ , $ Var(Traj) $, $ Var(E) $ have?  
% That question will be answered in Section \ref{sec:ComponentImpacts}, after we show % Surprisingly, 
% the three variances typically reported are sufficient to estimate variabilities of $ Z $, $ Traj $, and $ E $.  
\subsection{Plausibility}
Intuitively, variability at the first and milestone visits and of Change may contain all the information we need to separate the $ E $, $ Traj $, and $ Z $ variability components.  
Baseline measurements at the beginning of the study (visit 1) have the random intercept $ Z $ and measurement error $ E $ components but no random trajectory component $ Traj $ % $ \mu $ 
since patients have not been treated yet. 
Change, or \textit{change-from-baseline}, measurement at the milestone visit minus measurement at the first visit, has the random trajectory $ Traj $ % with expectation $ \mu $ 
and the measurement error $ E $ components (two copies of $ E $ in fact) but not the random intercept component since that has been subtracted off.  
Measurements at the end of the study (visit $ m $) have all three random components.  
So, assuming variability of measurement errors is constant across the visits, it seems plausible that $ Var(Y^{[1]}_{i}) $, $ Var(Y^{[m]}_{i}) $, and $ Var(Y^{[\mathrm{change}]}_{i}) $ are sufficient for us to recover the $ E $, $ Traj $, and $ Z $ variability components.

\subsection{Components of the ETZ decomposition}\label{sec:ETZDecomposition}

Our discussion starts without assuming variability of measurement error at visit $ m $ is the same as variability of measurement error at visit $ 1 $, or that the random intercept $ Z $ is independent of the random trajectory $ Traj $.  
As the decomposition proceeds, these two assumptions will be explicitly added as needed.  

With such a conceptualization, outcomes of patient $ r $ given $ Rx $ and patient $ s $ given $ C $ at visit 1 and visit $ m $ are represented as 
\begin{eqnarray}
	Y^{Rx[1]}_{r} & = & Z_{r} + E^{[1]}_{r} \label{eq:Ybefore}\\
	% Y^{Rx[m]}_{r} & = & Z_{r} + \mu^{[m]r}_{Rx} + E^{[m]}_{r} \\
	Y^{Rx[m]}_{r} & = & Z_{r} + Traj^{Rx[m]}_{r} + E^{[m]}_{r} \label{eq:Yafter}\\
	Y^{C[1]}_{s}  & = & Z_{s} + E^{[1]}_{s} \nonumber\\
	% Y^{C[m]}_{s}  & = & Z_{s} + \mu^{[m]s}_{C} + E^{[m]}_{s} \nonumber
	Y^{C[m]}_{s}  & = & Z_{s} + Traj^{C[m]}_{s} + E^{[m]}_{s}. \nonumber
\end{eqnarray}
Implicitly, the fixed effects in this model are 
\begin{eqnarray}
	\mathbbm{E}(Z_{r}) = \alpha_{Rx}  =  \mathbbm{E}(Z_{s})  & = & \alpha_{C} \label{eq:EqualAlpha}\\
	% \mathbbm{E}(Traj^{Rx[m]}_{r}) = \mu^{[m]r}_{Rx} & \mathrm{and} & \mathbbm{E}(Traj^{C[m]}_{s}) = \mu^{[m]s}_{C} 
	\mathbbm{E}(Traj^{Rx[m]}_{r}) & = & \mu^{Rx[m]}_{r} \\
	\mathbbm{E}(Traj^{C[m]}_{s}) & = & \mu^{C[m]}_{s} 
\end{eqnarray}
with $ \mathbbm{E}(Z_{r}) = \mathbbm{E}(Z_{s}) $ because patients are randomly assigned to $ Rx $ and $ C $.  
% Variances and covariances have the means (expected values) subtracted off, for both Z and Traj, jch 26 Mar 2022
% So $ Cov (Y^{Rx[1]}_{r}, Y^{Rx[m]}_{r}) = Var (Z_{r}) $, if $ Z $ and $ Traj $ are independent % jch 26 Mar 2022

Assuming variabilities are the same under $ Rx $ and $ C $, for patient $ r $ given treatment $ Trt $ which can be $ Rx $ or $ C $,
% \begin{align}
	% Var(Y^{Trt[1]}_{r}) & = & Var(\mathrm{random~intercept}~Z_{r})+ Var(\mathrm{visit ~1~measurement~error~} E^{[1]}_{r}) %\nonumber\\ & ~ &
	% \label{eq:Varvisit1} \\
	% Var(Y^{Trt[m]}_{r}) & = & Var(\mathrm{random~intercept}~Z_{r}) + Var(\mathrm{visit}~m~\mathrm{measurement~error}~E^{[m]}_{r}) + \nonumber\\
	% & & 2 ~ Cov(Z_{r} , Traj^{Trt[m]}_{r} ) + \nonumber \\	% if Z and Traj are dependent jch 26 Mar 2022
	% & & Var(\mathrm{random~trajectory}~Traj^{Trt[m]}_{r}) \label{eq:VarvisitM}
	% \end{align}
\begin{eqnarray}
	Var(Y^{Trt[1]}_{r}) & = & Var(Z_{r})+ Var(E^{[1]}_{r}) %\nonumber\\ & ~ &
	\label{eq:Varvisit1} \\
	Var(Y^{Trt[m]}_{r}) & = & Var(Z_{r}) + Var(E^{[m]}_{r}) \label{eq:VarvisitM}\\
	& & +2 ~ Cov(Z_{r} , Traj^{Trt[m]}_{r} )\nonumber \\	% if Z and Traj are dependent jch 26 Mar 2022
	& & + Var(Traj^{Trt[m]}_{r}), \nonumber
\end{eqnarray}
where $Z_{r}$ is the random intercept, $E^{[1]}_{r}$ is the random visit 1 measurement error, $E^{[m]}_{r}$ is the random visit $ m $ measurement error and $ Traj^{Trt[m]}_{r} $ is the random trajectory.

% For example, if % intercept = 0 and 
% the response profile is linear, then 
% $ Var(Traj^{Trt[m]}_{r}) % ~\mu^{[m]r}_{i}) $ 
% =  t^2_{m}Var(\mathrm{slope}_{Trt}) $ where $ t_{m} $ is time of the milestone visit.  
Note  
\begin{eqnarray*}
	Var(Z_{r}) + Cov(Z_{r} , Traj^{Trt[m]}_{r} )	% if Z and Traj are dependent jch 26 Mar 2022
	& = & Cov(Y^{Trt[m]}_{r}, Y^{Trt[1]}_{r}) 
\end{eqnarray*}
and, at this point, we have five unknown ETZ variability components: 
\begin{eqnarray*}
	Var(Z_{r}), Cov( Z_{r} , Traj^{Trt[m]}_{r} ), Var(Traj^{Trt[m]}_{r}), Var(E^{[1]}_{r}),\\
	\mathrm{~and~} Var(E^{[m]}_{r}).
\end{eqnarray*}
% $$ Var(Z_{r}), Cov( Z_{r} , Traj^{Trt[m]}_{r} ), Var(Traj^{Trt[m]}_{r}), Var(E^{[1]}_{r}),\\\mathrm{~and~} Var(E^{[m]}_{r}). $$  

\textit{Change}, or \textit{change-from-baseline}, is measured by subtracting the baseline (visit 1) measurement from the milestone (visit $ m $) measurement, so it does not have a random intercept component.  
\textit{Change} % (visit $ m $ outcome $ - $ visit 1 outcome) 
of patient $ r $ given treatment $ i $ is then 
% $ Rx $ and patient $ s $ given $ C $ are then 
\begin{eqnarray*}
	% Y^{[\mathrm{change}]r}_{Rx} = \left[ \mu^{[m]r}_{Rx} % - \mu^{[1]}_{Rx} 
	Y^{Trt[\mathrm{change}]}_{r} = \left[ Traj^{Trt[m]}_{r} % - Traj^{[1]}_{Rx} 
	\right] + \left[ E^{[m]}_{r} - E^{[1]}_{r} \right] 
	%	Y^{[\mathrm{change}]r}_{Rx} = \left[ Traj^{Rx[m]}_{r} % - Traj^{[1]}_{Rx} 
	%	\right] + \left[ E^{[m]}_{r} - E^{[1]}_{r} \right] \\
	% Y^{[\mathrm{change}]s}_{C} = \left[ \mu^{[m]s}_{C} % - \mu^{[1]}_{C}
	%	Y^{[\mathrm{change}]s}_{C} = \left[ Traj^{C[m]}_{s} % - Traj^{[1]}_{C}
	%	\right] + \left[ E^{[m]}_{s} - E^{[1]}_{s} \right] \nonumber
\end{eqnarray*}
so  
% \begin{eqnarray}
	% Var(Y^{Trt[\mathrm{change}]}_{r}) & = & Var(\mathrm{visit~1~measurement~error~} E^{[1]}_{r}) \nonumber\\
	% % Var(\mathrm{random~trajectory} ~\mu^{[m]r}_{i}) + \nonumber \\ & &
	% & & + Var(\mathrm{random~trajectory} ~Traj^{Trt[m]}_{r}) + \label{eq:VarChange}\\ 
	% & & + Var(\mathrm{visit}~m~\mathrm{measurement~error~} E^{[m]}_{r}) \nonumber
	% \end{eqnarray}
\begin{eqnarray}
	Var(Y^{Trt[\mathrm{change}]}_{r}) & = & Var(E^{[1]}_{r})\label{eq:VarChange} \\
	% Var(\mathrm{random~trajectory} ~\mu^{[m]r}_{i}) + \nonumber \\ & &
	& & + Var(Traj^{Trt[m]}_{r})\nonumber\\ 
	& & + Var(E^{[m]}_{r}) \nonumber
\end{eqnarray}
which leads to  
%\begin{eqnarray*}
%	Var(Y^{Trt[1]}_{r}) + Var(Y^{Trt[m]}_{r}) - Var(Y^{Trt[\mathrm{change}]}_{r}) 
%%	& = & 2 Cov(Y^{Trt[m]}_{r}, Y^{Trt[1]}_{r}) \\
%%	Var(Y^{Trt[1]}_{r}) + Var(Y^{Trt[m]}_{r}) - Var(Y^{Trt[\mathrm{change}]}_{r}) & = & 2 Cov(Y^{Rx[m]}_{r}, Y^{Rx[1]}_{r}) \\
%	& = & 2Var(\mathrm{random~intercept}~Z_{r})
%	+ \\ & & 2 Cov(Z_{r} , Traj^{Trt[m]}_{r} )	% if Z and Traj are dependent jch 26 Mar 2022
%\end{eqnarray*}
%\begin{small}
\begin{align}
	& ~ Var(Z_{r}) + Cov(Z_{r} , Traj^{Trt[m]}_{r})\nonumber\\	% if Z and Traj are dependent jch 26 Mar 2022
	& = 
	\dfrac{Var(Y^{Trt[m]}_{r}) + Var(Y^{Trt[1]}_{r}) - Var(Y^{Trt[\mathrm{change}]}_{r})}{2}  \label{eq:VarZest}
	\\ & = Cov(Y^{Trt[m]}_{r}, Y^{Trt[1]}_{r}) \label{eq:Covvisits1m}
\end{align}
%\end{small}
with the number of unknown ETZ variability components remaining five.  
% jch 20Apr2025
(Note that since estimates of the three variances in (\ref{eq:VarZest}) are routinely included as summary statistics in reporting RCT outcomes, $Cov(Y^{Trt[m]}_{r}, Y^{Trt[1]}_{r})$ in (\ref{eq:Covvisits1m}) can be estimated without access to patient level data.)  
Making the assumption $ Var(E^{[1]}_{r}) = Var(E^{[m]}_{r}) $, reducing the number of unknowns to four, gives 
\begin{eqnarray}
	& ~ &Var(Traj^{Trt[m]}_{r}) + 2 Cov( Z_{r} , Traj^{Trt[m]}_{r} ) \nonumber\\% if Z and Traj are dependent jch 26 Mar 2022
	& = & Var(Y^{Trt[m]}_{r}) - Var(Y^{Trt[1]}_{r}).\label{eq:VarTrajEst}
\end{eqnarray}

If we further make the assumption that $ Z $ and $ Traj $ are independent, so that $ Cov(Z_{r} , Traj^{Trt[m]}_{r} ) = 0 $, reducing the number of unknown variability components to three, then we can estimate $ Var(Z_{r}) $ from (\ref{eq:VarZest}), and estimate $ Var(Traj^{Trt[m]}_{r}) $ from (\ref{eq:VarTrajEst}).  
In turn, we can estimate variance of measurement error as 
\begin{eqnarray}
	Var(E^{[1]}_{r}) & = & Var(Y^{Trt[1]}_{r}) - Var(Z_{r}) \label{eq:VarEest}\\
	& = & Var(E^{[m]}_{r}). \nonumber % \\
	%	Var(E^{[m]}_{r}) & = & Var(Y^{Rx[m]}_{r}) - Var(\mu^{[m]}_{Rx}) - Var(Z_{r})
\end{eqnarray}
%\begin{eqnarray*}
%%	Var(Y^{[\mathrm{change}]r}_{Rx}) & = & Var(Y^{Rx[m]}_{r}) + Var(Y^{Rx[1]}_{r}) - 2 Cov(Y^{Rx[m]}_{r}, Y^{Rx[1]}_{r}) \\
%%	Var(Y^{[\mathrm{change}]r}_{Rx}) & = & Var(Y^{Rx[m]}_{r}) + Var(Y^{Rx[1]}_{r}) - 2 Var(Z_{r}) \\
%	Var(Z) & = & 
%	\left[ Var(Y^{Rx[m]}_{r}) + Var(Y^{Rx[1]}_{r}) - Var(Y^{[\mathrm{change}]r}_{Rx})\right] / 2 \\
%	 & = & Cov(Y^{Rx[m]}_{r}, Y^{Rx[1]}_{r}) \\
%	Var(Traj^{Trt[m]}_{r}) & = & Var(Y^{Rx[m]}_{r}) - Var(Y^{Rx[1]}_{r}) \\
%	Var(E^{[1]}_{r}) & = & Var(Y^{Rx[1]}_{r}) - Var(Z_{r}) \\
%				& = & Var(E^{[m]}_{r}) % \\
%%	Var(E^{[m]}_{r}) & = & Var(Y^{Rx[m]}_{r}) - Var(\mu^{[m]}_{Rx}) - Var(Z_{r})
%\end{eqnarray*}
%The EZ variance component decomposition comes from noticing 
%\begin{eqnarray*}
%	Var(Y^{[\mathrm{change}]r}_{Rx}) & = & Var(Y^{Rx[m]}_{r}) + Var(Y^{Rx[1]}_{r}) - 2 Cov(Y^{Rx[m]}_{r}, Y^{Rx[1]}_{r}) \\
%	Var(Y^{[\mathrm{change}]r}_{Rx}) & = & Var(Y^{Rx[m]}_{r}) + Var(Y^{Rx[1]}_{r}) - 2 Var(Z_{r}) \\
%	Var(Z) & = & \left[ Var(Y^{Rx[m]}_{r}) + Var(Y^{Rx[1]}_{r}) - Var(Y^{[\mathrm{change}]r}_{Rx})\right] / 2 \\
%	Var(E^{[1]}_{r}) & = & Var(Y^{Rx[1]}_{r}) - Var(Z_{r}) \\
%	Var(E^{[m]}_{r}) & = & Var(Y^{Rx[m]}_{r}) - Var(\mu^{[m]}_{Rx}) - Var(Z_{r})
%\end{eqnarray*}

\begin{figure}[hbtp]
	\centering
	\includegraphics[width=0.8\linewidth]{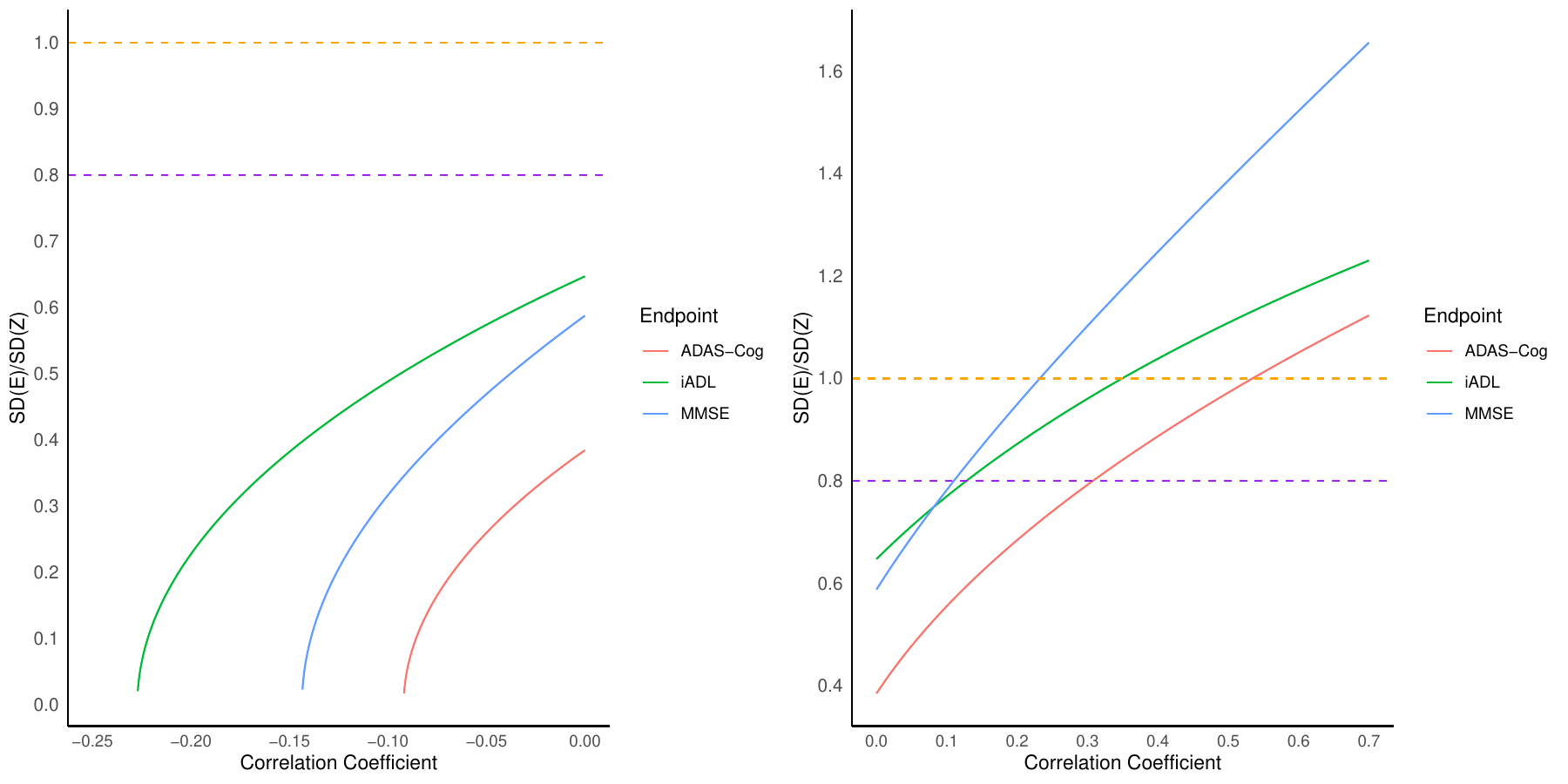}
	\caption{Ratio SD($ E $)/SD($ Z $) of standard deviations across different clinical endpoints (ADAS-Cog, iADL, and MMSE) as functions of the correlation between $ Z $ and $ Traj $, using EXPEDITION3 as an example. The left panel shows 
		if $ Z $ and $ Traj $ are sufficiently negatively correlated, then ETZ modeling can result in $Var( E ) < 0$ clearly indicating the independence assumption does not hold.  The right panel shows the extent to which positive correlation between $ Z $ and $ Traj $ can cause ETZ modeling with the independence assumption to miss the signal that $Var( E ) > Var( Z )$.  \textcolor{purple}{Purple} dashed line indicates when SD ($E$) reaches 80\% of SD ($Z$), while \textcolor{orange}{orange} dashed line indicates SD ($E$) = SD ($Z$).}
	\label{fig:NegPosCorrPlot}
\end{figure}

Figure \ref{fig:NegPosCorrPlot} examines limitations of current ETZ modeling with the independence assumption.   One potential issue is, if $ Z $ and $ Traj $ are in fact positively correlated, then from (\ref{eq:Covvisits1m}), (\ref{eq:VarTrajEst}), and (\ref{eq:VarEest}), estimates of $ Var(Z_{r}) $ and $ Var(Traj^{Trt[m]}_{r}) $ would be higher than they ought to be, while the estimate for $ Var(E^{[1]}_{r}) = Var(E^{[m]}_{r}) $ would be lower than it ought to be.  Then we might miss the signal that $Var( E ) > Var( Z )$, in which case study result is unreliable because within-patient measurement variability dominates between-patient variability.  The right panel of Figure \ref{fig:NegPosCorrPlot} shows how one can make a plot to assess how large the correlation needs to be for this signal to be missed.
Another potential issue is, if $ Z $ and $ Traj $ are sufficiently negatively correlated, then as shown in the left panel of Figure \ref{fig:NegPosCorrPlot}, ETZ modeling can result in $Var( E ) < 0$ which is nonsensical.  But we do not expect negative correlation to be common in practice, as that corresponds to patients at late stage of the disease having more favorable trajectories.

\subsection{Uncertainty reduced by \textit{counterfactual} uncertainty quantification}\label{sec:CUQ}

We now show exactly how \textit{Counterfactual Uncertainty Quantification} (CUQ), the goal of our research, is achieved by the ETZ modeling principle.  
% The two steps of \textit{counterfactual} uncertainty quantification, baselining and self-controlling, both reduce uncertainty from factual uncertainty qualification.
% Consequently, the number of patients enrolled in clinical trials can be reduced because sample size and variability are inversely proportional.

\paragraph*{\textbf{Narrowing patient entry criterion is ineffective}}
% We first demonstrate the intuition for uncertainty reduced by baselining using information available in FDA's review of Trulicity\textsuperscript{\textregistered} for treating patients with type 2 diabetes.
There is the thinking that narrowing the patient entry criterion makes patients more homogeneous and thereby makes clinical trials more likely to succeed. 
This turns out to not be the case if treatment effects are measured in terms of \textit{change-from-baseline}, because   
(\ref{eq:VarChange}) in the ETZ decomposition shows SD(Change) does not involve patients' baseline measurements $ Z $.  
% We demonstrate this using information available in FDA's review of Trulicity\textsuperscript{\textregistered} for treating patients with type 2 diabetes.

Trulicity\textsuperscript{\textregistered} (compound name dulaglutide) is a once-weekly injectable prescription medicine to improve blood sugar (glucose) in adults with type 2 diabetes (mellitus), as measured by control of haemoglobin A1c (HbA1c).  
Biologic License Application 125469 % which led to the approval of 
for Trulicity/dulaglutide had five studies.  
Patient entry criteria, SD(Basline), and SD(Change) for studies GBDA and GBDC, reported on pages 6 and 7 in \cite{TrulicityFDAsumaryReview(2014)} and on pages 58 and 61 of \cite{TrulicityFDAstatisticalReview(2014)}, are reproduced in Tables \ref{table:TrulicityRewindGBDA} and \ref{table:TrulicityRewindGBDC}.   

\begin{table}[hbtp]
	\centering
	\resizebox{\linewidth}{!}{
		\begin{tabular}{cccc}
			\toprule
			HbA1c & Dulaglutide + Metformin + Pioglitazone & Placebo & Exenatide \\
			\midrule
			\midrule
			$ n $ & 279 & 141 & 276 \\
			\midrule
			SD(Baseline) & 1.3 & 1.3 & 1.3 \\
			\midrule
			SD(Change) & 0.98 & 0.97 & 1.02 \\
			\bottomrule
	\end{tabular}}
	\caption{GBDA sub-study: Dulaglutide + Metformin + Pioglitazone vs. Placebo and Exenatide with baseline HbA1c $ \in [7, 11] $}\label{table:TrulicityRewindGBDA}
\end{table}

\begin{table}[hbtp]
	\centering
	%\resizebox{\linewidth}{!}{
		\begin{tabular}{ccc}
			\toprule
			HbA1c & Dulaglutide & Metformin \\
			\midrule
			\midrule
			$ n $ & 269 & 268 \\
			\midrule
			SD(Baseline) & 0.9 & 0.8 \\
			\midrule
			SD(Change) & 1.00 & 0.98 \\
			\bottomrule
		\end{tabular}%}
	\caption{GBDC sub-study: Dulaglutide vs. Metformin with baseline HbA1c $ \in [6.5, 9.5] $}\label{table:TrulicityRewindGBDC}
\end{table}

As can be seen, narrowing the HbA1c baseline entry criterion from [7, 11] in GBDA to [6.5, 9.5] in GBDC decreases SD(baseline) of HbA1c from 1.3 to within [0.8, 0.9], but hardly affects SD(Change), which remains steady within [0.97, 1.02] for both studies.  
So decreasing variability of intercept $ Z $ may not reduce CUQ.  

\paragraph*{\textbf{Baselining significantly reduces uncertainty}}
ETZ pinpoints when baselining effectively reduces uncertainty.  

We show below that baselining effectively replaces the patient-to-patient variability $ Var (Z) $ by the measurement error variability $ Var (E) $ and, provided that $ Var (Z) > Var (E) $, baselining is beneficial.  

\begin{figure}[hbtp]
	\centering
	\includegraphics[scale = 0.6]{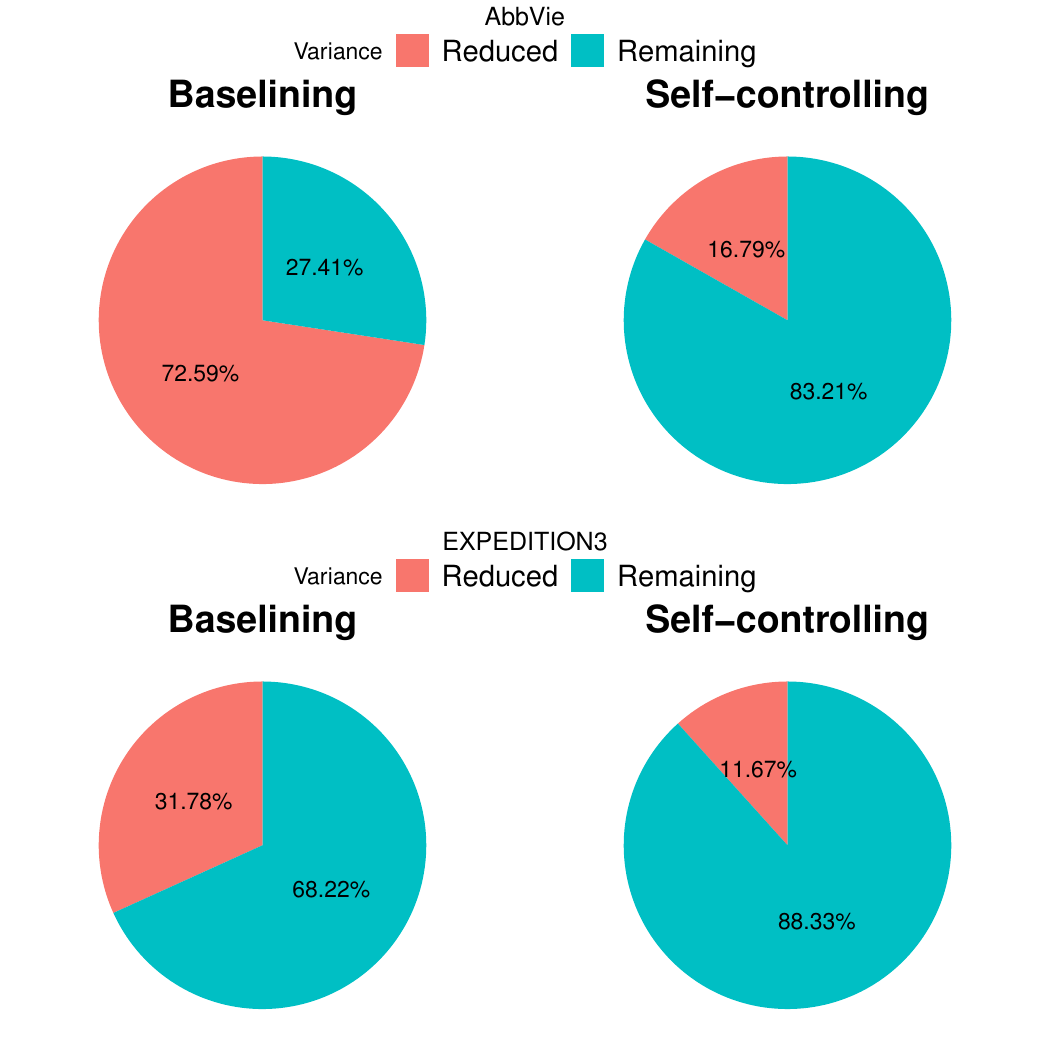}
	\caption{\textit{Counterfactual} uncertainty quantification can reduce the quantified uncertainty in efficacy estimation. The pie charts on the left show the proportion of total variance reduced by \textit{baselining} while the pie charts on the right show the additional proportion reduced by \textit{counterfactual} (instead of factual) calculations. The top two pie charts analyze data from the AbbVie study, while the bottom two pie charts analyze data from the EXPEDITION3 study. Overall, \textit{Counterfactual Uncertainty Quantification} reduces 77.19\% and 39.74\% of the variances in the AbbVie study and the EXPEDITION3 study, respectively. The detailed calculation has been demonstrated in Appendix \ref{appC}.}
	\label{fig:CUQexamples}
\end{figure}

Whichever treatment patient $i$ is assigned to, expressions for $Var (Y_i^{[m]})$ and $Var (Y_i^{[change]})$ are 
\begin{eqnarray}
	Var (Y_i^{[m]}) & = & Var (Z_i) + Var (Traj_i) + Var (E_i^{[m]})\\
	Var (Y_i^{[1]}) & = & Var (Z_i) + Var (E_i^{[1]})\\
	Var (Y_i^{[change]}) & = & Var (Y_i^{[m]} - Y_i^{[1]})\nonumber\\
	& = & Var (Z_i + Traj_i + E_i^{[m]} - Z_i - E_i^{[1]})\nonumber\\
	& = & Var ( Traj_i + E_i^{[m]} - E_i^{[1]})\nonumber\\
	& = & Var (Traj) + 2\times Var (E). \label{eq: ChangeDef}
\end{eqnarray}
with (\ref{eq: ChangeDef}) showing the expression for the variance of \textit{change-from-baseline}.  
So the difference between $Var (Y_i^{[m]})$ and $Var (Y_i^{[change]})$ is 
\begin{eqnarray}
	& &Var (Y_i^{[m]}) - Var (Y_i^{[change]}) \nonumber\\
	& = & Var (Z) - Var (E)\label{eq: DiffMileChange}
\end{eqnarray}
and $Var (Y_i^{[change]}) < Var (Y_i^{[m]})$ when $Var (E) < Var (Z)$.  
(Note that if measurement variability is larger than patient-to-patient variability, $ Var (E) > Var (Z) $, then it is hopeless for that clinical trial to be informative and a different outcome measure should be chosen.)  
Figure \ref{fig:CUQexamples} in Section \ref{sec:PieCharts} shows that {baselining can reduce total variability by 73\%} in an example data set.  

\paragraph*{\textbf{Self-controlling CUQ further reduces uncertainty}}
Beyond baselining, \textit{self-controlling} CUQ further reduces uncertainty by two copies of the measurement error variance, using ETZ to quantify uncertainty counterfactually instead of factually.

Factual uncertainty quantification considers a randomly selected pair of patients with patient $r$ given $Rx$ and patient $s$ given $C$.  
With $Y_r^{Rx[change]}$ and $Y_s^{C[change]}$ denoting their \textit{change-from-baseline} outcomes, factual efficacy $Y^{f}$ is estimated by 
$$Y^{f} = Y_r^{Rx[change]} - Y_s^{C[change]}.$$

Self-controlling  uncertainty quantification considers the difference between potential \textit{change-from-baseline} of the same patient $i$ receiving both $Rx$ and $C$.
With $Y_i^{[change]}(Rx)$ and $Y_i^{[change]}(C)$ denoting the potential \textit{change-from-baseline} outcomes for patient $i$, 
truly  \textit{counterfactual} efficacy $Y^{cf}$ is estimated by
$$Y^{cf} = Y_i^{[change]}(Rx) - Y_i^{[change]}(C).$$

ETZ assumes $Var (E)$ remains constant and $Var (Traj) = Var (Traj^{Rx}) = Var (Traj^{C})$.
Moreover, $ E $ and $ Traj $ are independent because how patients respond to treatment is irrelevant to measurement error.

Hence, factual variability is 
\begin{eqnarray}
	Var (Y^{f}) & = & Var (Y_r^{Rx[change]} - Y_s^{C[change]})\nonumber\\
	& = & Var (Traj_r^{Rx} + E_r^{[1]}+ E_r^{Rx[m]}  \nonumber\\
	& & - Traj_s^{C} - E_s^{[1]} - E_s^{C[m]})  \nonumber\\\label{eq: EBaseline}
	& = & 2\times Var (Traj) + 4 \times Var (E),
\end{eqnarray}
where $Traj_r^{Rx}$ and $Traj_s^{C}$ are independent because patients $r$ and $s$ are different patients.  
On the other hand, counterfactual variability is 
\begin{eqnarray}
	Var (Y^{cf}) & = & Var (Y_i^{Rx[change]} - Y_i^{C[change]})\nonumber\\
	& = & Var (Traj_i^{Rx} + E_i^{[1]}+ E_i^{Rx[m]}\nonumber\\
	& &- Traj_i^{C} - E_i^{[1]} - E_i^{C[m]})\nonumber\\
	& = & Var (Traj_i^{Rx} + E_i^{Rx[m]} - Traj_i^{C} - E_i^{C[m]})\nonumber\\
	& = & 2\times Var (Traj) + 2 \times Var (E) \nonumber \\
	& ~ &- 2 \times Cov (Traj_i^{Rx},Traj_i^{C}).\label{eq: EBaselineCF}
\end{eqnarray}
Thus, 
\begin{eqnarray}
	& & Var (Y^{f}) - Var (Y^{cf})\nonumber \\
	& = & 2 \times Var (E) + 2 \times Cov (Traj_i^{Rx},Traj_i^{C}).
\end{eqnarray}
where $Cov (Traj_i^{Rx},Traj_i^{C})$ is non-negative because trajectories from the same patient $i$ to $ Rx $ and $ C $ would be independent or positively correlated as explained below.

Biologically, if $ C $ is a placebo while $ Rx $ is a targeted therapy, then one can expect the trajectories to be independent.  
If both $ Rx $ and $ C $ are targeted therapies but with entirely different targets, for example $Rx$ is GLP-1 while $C$ is metformin in treating T2DM patients, then one would expect the trajectories to be independent.  
% Simvastatin and losartan are both used to improve cardiovascular health.
%But simvastatin works by inhibiting HMG-CoA reductase, while losartan works by blocking the angiotensin II receptors (AT1 receptors).

Trajectories can expect to be positively correlated when $ Rx $ and $ C $ have similar mechanism of action or are upstream and downstream of the same cascade or pathway.
For example, $Rx$ and $C$ 
% like Chlorpromazine and Olanzapine 
for treating schizophrenia may both be dopamine receptor antagonists.  
%and both work by blocking dopamine receptors to reduce the effects of dopamine and alleviate psychotic symptoms.
For treatment of Alzheimer's disease, $C$ might target $\beta$-amyloid plaque (upstream) while $Rx$ targets tau-tangle (downstream).  
%the accumulation of $\beta$-amyloid may be an upstream event that triggers the formation of tau tangles.
%Solanezumab and LMTX are both medicines for Alzheimer's disease.
%But solanezumab targets at $\beta$-amyloid while LMTX targets at tau tangles.

So self-controlling removes \textbf{at least} two copies of the measurement error variance, and our claim of CUQ reducing $2 \times Var (E)$ may even be conservative.  
Figure \ref{fig:CUQexamples} in Section \ref{sec:PieCharts} shows that {self-controlling CUQ can reduce an additional 17\% of uncertainty beyond baselining} in a real example.  
%In equation (\ref{eq: EBaseline}), $E_r^{[1]Rx}$ and $E_s^{[1]C}$ can be written as $E_r^{[1]}$ because no patient is assigned to any treatment at baseline.
%In equation (\ref{eq: EBaselineCF}), the measurement error for patient $i$ at first visit is offset by subtraction.

\subsection{Real data examples of CUQ}\label{sec:PieCharts}
In this section, we show the proportion of variance reduced by \textit{counterfactual} uncertainty quantification using two real clinical datasets, the AbbVie dataset and the EXPEDITION3 dataset.

\cite{schnell2017subgroup} % Schnell et al. (2017)
described an Alzheimer Disease study from AbbVie which compared three doses of a new treatment (doses 1, 2, and 3) with a negative control (dose 0, a placebo).  
% There was an active control (dose 4) in the study as well.  
% Outcome measure was ADAS-cog11, and duration of this Phase 2 study was 24 weeks. 
The AbbVie data set is described in Section \ref{sec:IndCase}.

%Honig et al. (2018) 
\cite{honig2018trial} 
reported on EXPEDITION3, a double-blind, placebo-controlled, Phase 3 study testing solanezumab on patients with mild dementia due to Alzheimer’s disease.  
Among a total of 2129 patients enrolled, 1057 were assigned to receive solanezumab while 1072 were assigned to the placebo.
Score at the initial (Week 0) visit before randomization was considered \textit{baseline}.
Thereafter outcomes were measured at weeks 12, 28, 40, 52, 64, and 80. 
\textit{Change-from-baseline} is the difference between milestone outcome (at Week 80) and \textit{baseline} (at Week 0).
In EXPEDITION 3, the primary endpoint was the 14-item cognitive subscale of the Alzheimer’s Disease Assessment Scale (ADAS-cog14), while the key secondary endpoint was instrumental Activities of Daily Living (iADL).
EXPEDITION3 did not demonstrate efficacy in the primary endpoint ADAS-cog14 but showed promising result in the secondary endpoint iADL. 
We use iADL for illustration.

% %Combine the four pie charts?
% \begin{figure}
	%     \centering
	%     \includegraphics[scale = 0.4]{Figure/EXPEDITION3Example.eps}
	%     \caption{Caption}
	%     \label{fig:EXPEDITION3Example}
	% \end{figure}

\section{A broader % counterfactual estimation perspective}
framework}

A broader \textit{counterfactual} estimation framework includes both Real Human Counterfactual Estimation (RHCE) and Digital Twin Counterfactual Estimation (DTCE).
% Our methodology is self-controlling counterfactual estimation (SCCE), with baselining and variance reduction by ETZ modeling, 
% with each person serving as his/her own control.  % our strategy is applicable when
Our methodology is RHCE, baselining and self-controlling in the ETZ modeling principle with each human serving as their own control to reduce uncertainty quantification.  

Another strategy for \textit{counterfactual} estimation is to collect biomarker data on each patient, 
and train Digital Twin (DT) algorithms to predict each patient's outcome from their biomarker values.
% \begin{itemize}
%     \item train a $C$ outcome prediction algorithm $DT^{C}$ by learning from $C$-treated patient outcomes, 
%     \item train an $Rx$ outcome prediction algorithm $DT^{Rx}$ by learning from $Rx$-treated patient outcomes.  
% \end{itemize}
% Then create digital twin counterfactual outcomes: 
% \begin{itemize}
%     \item use $DT^{C}$ to predict the $C$ outcome of each $Rx$-treated patient, 
%     \item use $DT^{Rx}$ to predict the $Rx$ outcome of each $C$-treated patient.  
% \end{itemize}
In Digital Twin Counterfactual Estimation (DTCE), population $Rx: C$ efficacy is estimated by averaging across the patients the difference of their $Rx$ outcome and $C$ outcome.  

%Conceptually, while RHCE uses each patient as his/her own control and then models treatment effects and variability components at the population level, DTCE models factual outcome to predict based on each patient's biomarker values his/her \textit{counterfactual} outcome.  
%Where RHCE and DTCE intercept is when the biomarker is \textit{baseline}, a special case giving deep insight into what data and efficacy frameworks are required to allow for \textit{unbiased} estimation of true treatment effects.  

\begin{figure}[hbtp]
\centering
\includegraphics[scale = 0.4]{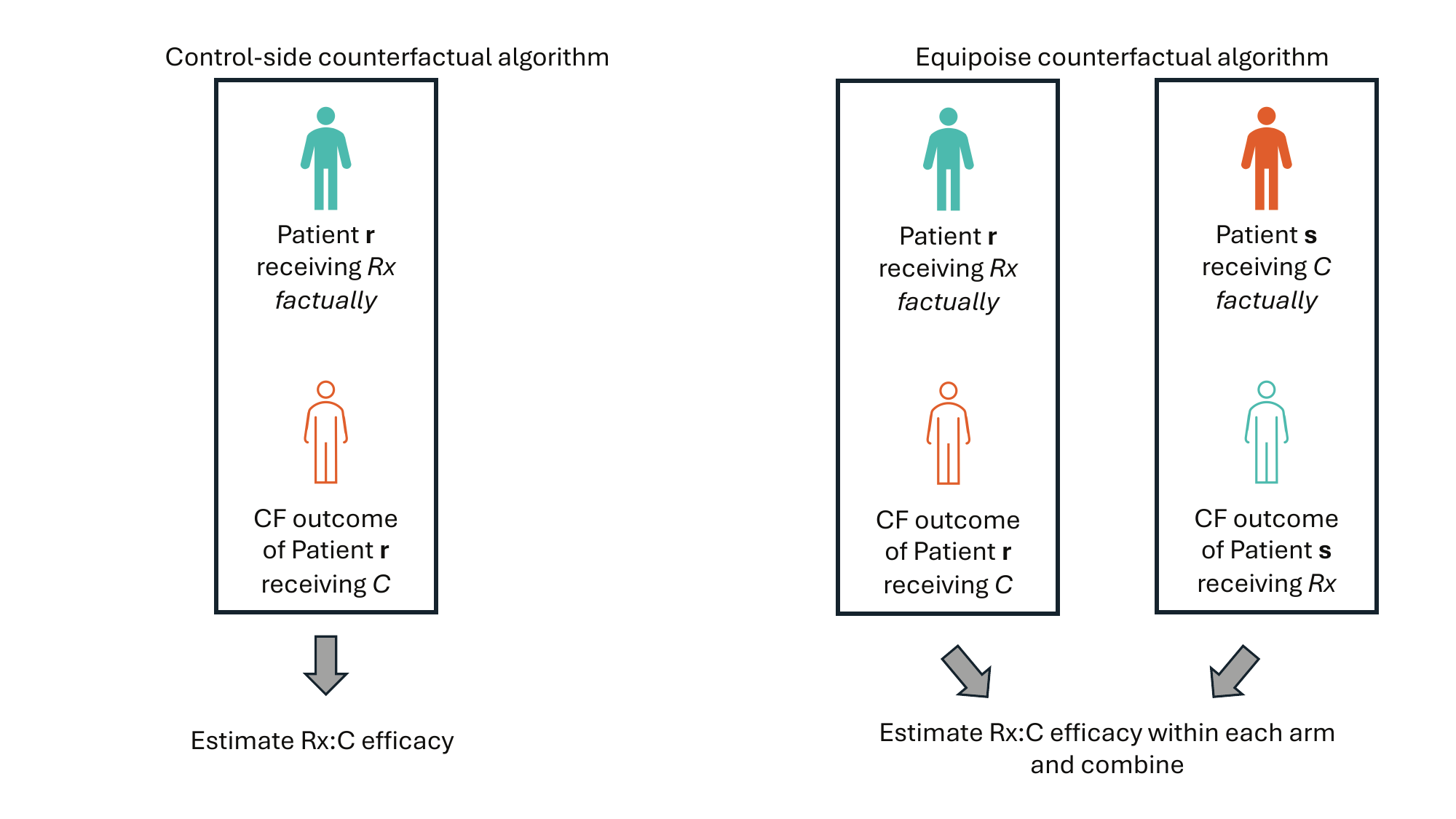}
\caption{\textit{Control-side} and \textit{Equipoise} counterfactual efficacy estimation.}
\label{fig:CFAlgorithm}
\end{figure}

\begin{figure}[hbtp]
\centering
\includegraphics[scale = 0.5]{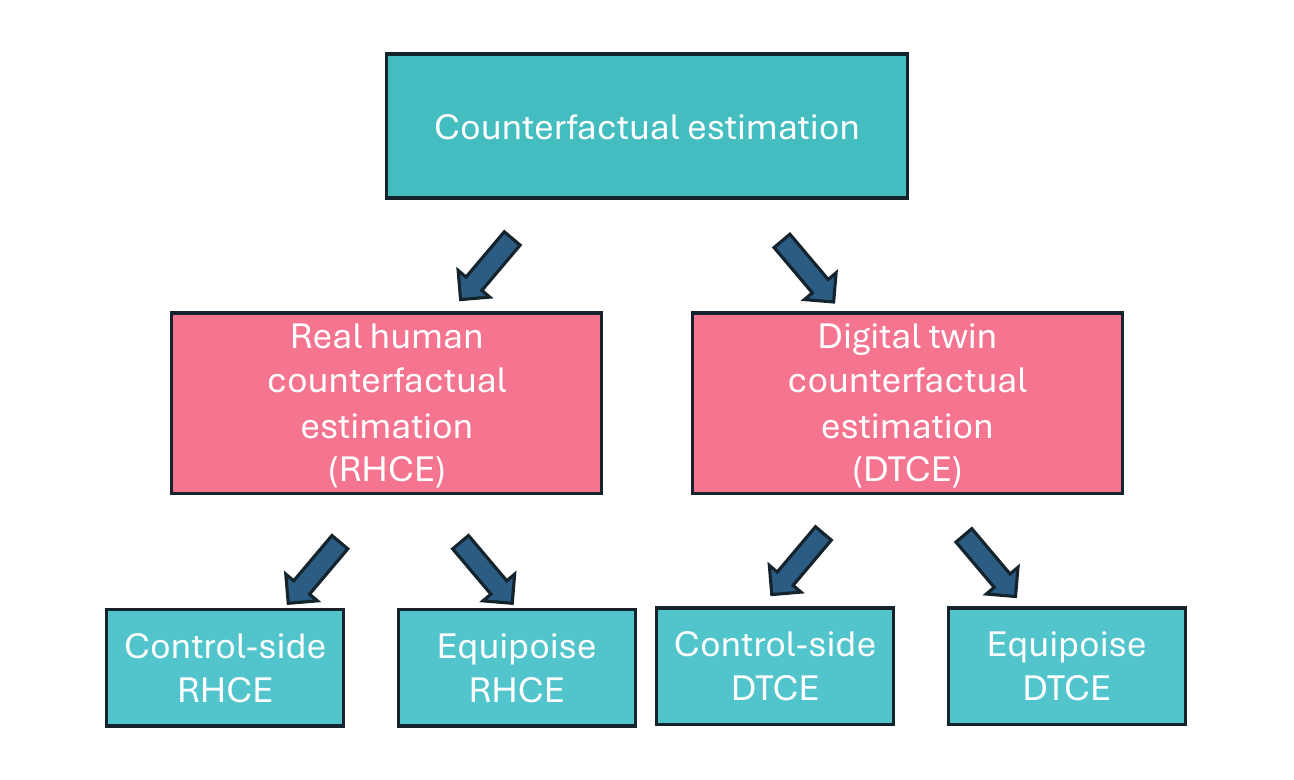}
\caption{\textit{Counterfactual} estimation includes Real Human Counterfactual Estimation (RHCE) and Digital Twin Counterfactual Estimation (DTCE), either can be control-side or equipoise.}
\label{fig:CFCategory}
\end{figure}

To explain potential biases that both RHCE and DTCE need to avoid, we separate \textit{Counterfactual} estimation as either on the  \textit{control-side} or \textit{equipoise}, as illustrated in Figure \ref{fig:CFAlgorithm}.
\textit{Control-side} counterfactual estimation, $CF_{control}$, takes only patients given $Rx$ and generates counterfactual $ C $ outcomes.  
\textit{Equipoise} counterfactual estimation, $CF_{equipoise}$, not only takes patients given $ Rx $ and generates counterfactual $ C $ outcomes but also generates counterfactual $ Rx $ outcomes for patients given $C$. 
While $CF_{equipoise}$ has been standard practice, $CF_{control}$ is more recent thinking, in the DTCE setting.  

In Section \ref{sec:Targetted} that follows, we define traditional medicine versus modern targeted therapies, and explain how they have different response profiles.  
% First, Figure \ref{fig:TradTargetCom} gives a visual preview that $CF_{control}$ may not even be unbiased with traditional medicine, while $CF_{equipoise}$ however is unbiased for traditional medicine, but both $CF_{control}$ and $CF_{equipoise}$ can be biased for targeted therapies, as will be shown in Section \ref{sec:CFEUnbiased}. 
First, Figure \ref{fig:TradTargetCom} gives a visual preview that $CF_{control}$ may not even be applicable to traditional medicine, while $CF_{equipoise}$ however is unbiased for traditional medicine, but both $CF_{control}$ and $CF_{equipoise}$ can be biased for targeted therapies, as will be shown in Section \ref{sec:CFEUnbiased}. 

\subsection{Genesis of baseline covariate adjustment}\label{sec:GenesisBaseline}
The purpose of including \textit{baseline} as a covariate in current MMRM analysis is to adjust for imbalance in the proportion of $Rx$ and $C$ treated patients in the data, to avoid bias.  
It is appropriate for traditional medicine with a \textit{baseline} which is prognostic of outcome \emph{irrespective of the therapy received} in the sense described on pages 19-22 of the %FDA-NIH (2016)
\cite{BEST(2016)} BEST document.%\cite{BEST(2016)}.  
% It is not for prediction purpose.  

If the data contains a substantial imbalance,  
%in the proportion of $Rx$ and $C$ treated patients as classified by a factor \textit{prognostic} of outcome, 
then a real example in \cite{Hsu(1992)GLM} %Hsu (1992)
showed the \emph{marginal} analysis implemented in the \textbf{Means} statement of SAS Proc GLM which does not adjust for the \textit{prognostic} factor as a covariate can give very biased $Rx: C$ efficacy estimates.  
% , in fact exhibiting  Simpson's paradox-like behavior in cases of extreme imbalance.  

Subsequently, starting with version 6.12 in 1996, SAS follows Chapter 7 of \cite{Hsu(1996)} %Hsu (1996) 
to provide unbiased multiple comparisons in the \textbf{LSmeans} statement, adjusting for covariate(s) presumed equally prognostic for $ Rx $ and $ C $ using Gauss-Markov least square means estimation to map the sample to a ``balanced'' population.  

This practice has served the analysis of traditional medicine well for decades.  
However, targeted therapies are becoming quite common, and there is a possibility that for some of them \textit{baseline} may be \textit{predictive} of outcome in the sense described on pages 35-40 of the FDA/NIH (2016) BEST document, %\cite{BEST(2016)}, 
in which case one has to be more cautious.  

\subsection{From traditional medicine to targeted therapy}\label{sec:Targetted}

\paragraph*{\textbf{Traditional medicine}}
Traditional medicine is ``one-drug-fits-all'', prescribing the same compound with no specific drug target to all patients for a particular indication.  
For example, label of olanzapine (Zyprexa\textsuperscript{\textregistered}), which is indicated for patients with schizophrenia, states ``The mechanism of action of olanzapine, as with other drugs having efficacy in schizophrenia, is \emph{unknown}'' (\cite{olanzapineLabel(2009)}). %.  

\paragraph*{\textbf{Targeted therapies}}
Targeted therapies are \textit{personalized} medicine, also called \textit{precision} medicine, targeting specific disease pathways.  

A monoclonal antibody, an amino acid chain (with compound suffix \textit{mab}) that binds to a specific \textit{drug target} is a targeted therapy.  
Immunotherapies for cancer such as 
% atezolizumab (Tecentriq\textsuperscript{\textregistered}), 
nivolumab (Opdivo\textsuperscript{\textregistered}) and pembrolizumab (Keytruda\textsuperscript{\textregistered}) are examples of targeted therapies.  
% boosting the immune T-cell response against cancer cells, which can shrink some tumors or slow their growth. 

A tyrosine kinase inhibitor, a small molecule with compound suffix \textit{tinib}, each intending to block a specific reaction within cells, is also a targeted therapy.  
Examples of \textit{tinibs} are ibrutinib (for chronic lymphocytic leukemia CLL) and baricitinib (for rheumatoid arthritis RA).  

Different patients receiving the same targeted therapy may derive differential benefits, because the extent to which the drug target presents itself in patients may differ by patient.  

%Explain the two types of medicine (maybe put earlier in independent assumption section?)
%Give a plot comparing traditional medicine and targeted therapy
\begin{figure}[hbtp]
\centering
\includegraphics[scale = 0.6]{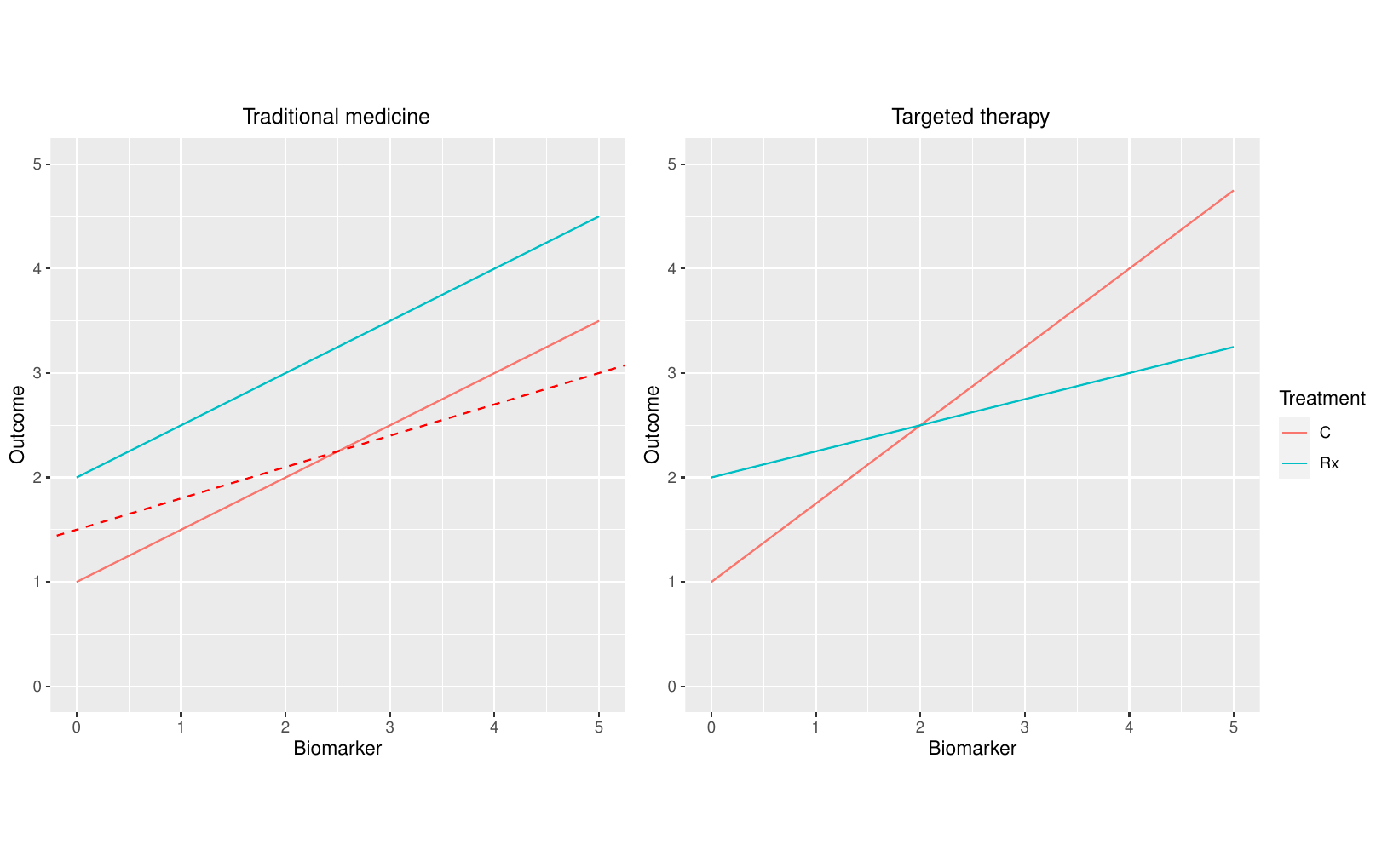}
\caption{Conceptual plots of response profiles for both traditional medicine and targeted therapy. The left plot shows a traditional medicine has parallel profiles while the right plot shows a targeted therapy may well have non-parallel profiles. The Red dashed line in the top plot previews a situation where the \emph{apparent} control-side counterfactual outcome profile is not parallel to the teal solid line, even for traditional medicine, in which case control-side counterfactual outcomes should \emph{not} be produced, as will be explained in Section \ref{sec:CFbiased}.}
\label{fig:TradTargetCom}
\end{figure}
%introduce the two models used
For both traditional medicine and targeted therapy, consider the simplest framework in which there are only two visits, \textit{baseline} and \textit{milestone}, and the true profiles are linear with random before treatment patient intercept $ Z $ independent of random milestone measurement error.   % $\varepsilon$ (parallel model) and $\epsilon$ (not parallel model).
Referring to the patterns shown in Figure \ref{fig:TradTargetCom}, we name the model used for traditional medicine as \textit{parallel traditional} medicine model and name the model used for targeted therapy as \textit{non-parallel targeted} therapy model.

A parallel traditional medicine model would be 
\begin{eqnarray}
Y^{Rx[m]}_{r} & = & \prescript{_Z}{}\theta_{} + \prescript{_Z}{}\eta^{Rx} + \prescript{_Z}{}\beta Z_{r} + \prescript{_Z}{}\varepsilon^{Rx[m]}_{r} \nonumber\\
Y^{C[m]}_{s} & = & \prescript{_Z}{}\theta_{} + \prescript{_Z}{}\eta^{C} + \prescript{_Z}{}\beta Z_{s} + \prescript{_Z}{}\varepsilon^{C[m]}_{s}, \label{mod:ParallelmodelZ}
\end{eqnarray}
where $\prescript{_Z}{}\theta_{}$ is the expected outcome at baseline visit before either treatment is given, $\prescript{_Z}{}\eta^{Rx}$ is the fixed $Rx$ treatment effect, $\prescript{_Z}{}\eta^{C}$ is the fixed $C$ treatment effect, $\prescript{_Z}{}\beta$ is the same coefficient for $Z$ shared by treatment $Rx$ and $C$, $\prescript{_Z}{}\varepsilon^{Rx[m]}_{r}$ and $\prescript{_Z}{}\varepsilon^{C[m]}_{s}$ are the measurement error for $Rx$ and $C$ respectively, so patients sicker at the beginning remain relatively sicker at the end irrespective of $ Rx $ or $ C $.  

For targeted therapy, a non-parallel targeted therapy model includes an additional interaction between $ Z $ and treatment $ Rx: C $ 
\begin{eqnarray}
Y^{Rx[m]}_{r} & = & \prescript{_Z}{}\alpha_{} + \prescript{_Z}{}\gamma^{Rx} + \prescript{_Z}{}\beta^{Rx}Z_{r} + \prescript{_Z}{}\epsilon^{Rx[m]}_{r} \nonumber\\
Y^{C[m]}_{s} & = & \prescript{_Z}{}\alpha_{} + \prescript{_Z}{}\gamma^{C} + \prescript{_Z}{}\beta^{C}Z_{s} + \prescript{_Z}{}\epsilon^{C[m]}_{s}, \label{mod:DTZmodel}
\end{eqnarray}
where $ \prescript{_Z}{}\alpha_{} $ is the expected outcome at baseline before either treatment is given, $\prescript{_Z}{}\gamma^{Rx}$ and $\prescript{_Z}{}\gamma^{C}$ are the fixed treatment effects for $Rx$ and $C$, $\prescript{_Z}{}\beta^{Rx}$ and $\prescript{_Z}{}\beta^{C}$ are \textit{different} coefficients for $ Z $ depending on whether the treatment is $ Rx $ or $ C $, $\prescript{_Z}{}\epsilon^{Rx[m]}_{r}$ and $\prescript{_Z}{}\epsilon^{C[m]}_{s}$ are the corresponding measurement errors,  with patients sicker at the beginning benefiting more from targeted theory $ Rx $ than from non-targeted therapy $ C $ expressed as $\prescript{_Z}{}\beta^{Rx}$ being a steeper slope than $\prescript{_Z}{}\beta^{C}$.  

Let $(B_1,B_2,\ldots,B_p)$ be a vector of $ p $ biomarkers and let $ B = g(B_1,B_2,\ldots,B_p) $ be a biomarker \emph{score} calculated to be on the same scale as the response (the primary endpoint), with $B$ potentially predictive of outcome.  

% Suppose in practice, for patients $r$ and $s$, $B_r = Z_{r}+\prescript{_B}{}E^{[1]}_{r}$ and $B_s = Z_{s}+\prescript{_B}{}E^{[1]}_{s}$ with measurement errors are used instead of their (not observable) $ Z $ values for prediction and ordinary regression is applied to the models (\ref{mod:ParallelmodelZ}) and (\ref{mod:DTZmodel}).
In practice, for patients $r$ and $s$, $B_r = Z_{r}+\prescript{_B}{}E^{[1]}_{r}$ and $B_s = Z_{s}+\prescript{_B}{}E^{[1]}_{s}$ with measurement errors are used instead of their (not observable) $ Z $ values for prediction and ordinary regression is applied to the models (\ref{mod:ParallelmodelZ}) and (\ref{mod:DTZmodel}).
\cite{burman2024digital} cites PROCOVA\textsuperscript{\texttrademark}
	as an example of Digital Twin, deriving a prognostic score from historical patient-level baseline data and using it as a covariate in ANCOVA modeling to boost trial efficiency.  
Appendix D of \cite{schuler2022increasing} presents the predictors PROCOVA used to generate the prognostic score.
Measurements of a number of the predictors, such as ADAS comprehension, CDR home and hobbies, and MMSE language, are subject to measurement error.

Then the parallel traditional medicine model becomes
\begin{eqnarray}
Y^{Rx[m]}_{r} & = & \prescript{_B}{}\theta_{} + \prescript{_B}{}\eta^{Rx} + \prescript{_B}{}\beta B_{r} + \prescript{_B}{}\varepsilon^{Rx[m]}_{r} \nonumber\\
Y^{C[m]}_{s} & = & \prescript{_B}{}\theta_{} + \prescript{_B}{}\eta^{C} + \prescript{_B}{}\beta B_{s} + \prescript{_B}{}\varepsilon^{C[m]}_{s} \label{mod:ParallelmodelB}
\end{eqnarray}
while the non-parallel targeted therapy model becomes
\begin{eqnarray}
Y^{Rx[m]}_{r} & = & \prescript{_B}{}\alpha_{} + \prescript{_B}{}\gamma^{Rx} + \prescript{_B}{}\beta^{Rx}B_{r} + \prescript{_B}{}\epsilon^{Rx[m]}_{r} \nonumber\\
Y^{C[m]}_{s} & = & \prescript{_B}{}\alpha_{} + \prescript{_B}{}\gamma^{C} + \prescript{_B}{}\beta^{C}B_{s} + \prescript{_B}{}\epsilon^{C[m]}_{s}\label{mod:DTWmodel}
% jch 21Mar2024 seems to me Z should be B = baseline
\end{eqnarray}
\noindent as if $B_r$ and $B_s$ were measured without errors.

\subsection{Noisy predictor causes attenuation}\label{sec:Attenuation}
%It is better that [4] comes before [1], because it can make people connect biomarker with measurement error with bias 
%The concept of "control side" and "equipoise" should be introduced before this section. (already done)

Regression modeling assumes predictors are measured without error, so what if some of $(B_1,B_2,\ldots,B_p)$ have measurement error, and $ B $ in turn has measurement error?  
For example, where as age and gender are presumably measured without error, \textit{baseline} will have measurement error.  
Turns out if $B$ has measurement error, then estimated $Rx$ and $C$ effects will have so-called \emph{attenuation}, and whether RHCE or DTCE is biased or unbiased for $Rx: C$ efficacy depends on how exactly that efficacy is defined, as will be shown in Section \ref{sec:CFbiased} and Section \ref{sec:CFEUnbiased}.  
We explain \textit{attenuation} in the context that a biomarker score $B$ for each patient is used in RHCE, with $B$ being \textit{baseline}.  

%The estimation of treatment effect is biased because of attenuation, but the bias will be offset when efficacy is measured at average of biomarker.
%Both \textit{control-side} and \textit{equipoise} \textit{counterfactual} estimation are considered in our discussion.

% To understand \textit{attenuation bias}, the simplest framework is where there are only two visits, \textit{baseline} and \textit{milestone} visits, and the true profiles are linear with random before treatment patient intercept $ Z $ independent of random milestone measurement error $\epsilon$:
% \begin{eqnarray}
% Y^{Rx[m]}_{r} & = & \prescript{_Z}{}\alpha_{} + \prescript{_Z}{}\gamma^{Rx} + \prescript{_Z}{}\beta^{Rx}Z_{r} + \prescript{_Z}{}\epsilon^{Rx[m]}_{r} \nonumber\\
% Y^{C[m]}_{s} & = & \prescript{_Z}{}\alpha_{} + \prescript{_Z}{}\gamma^{C} + \prescript{_Z}{}\beta^{C}Z_{s} + \prescript{_Z}{}\epsilon^{C[m]}_{s}. \label{mod:DTZmodel}
% \end{eqnarray}
% jch 2June2024 In this model, sicker patient may benefits more  
%Add the reference for the models here

%\paragraph*{\textbf{Attenuation}}%tentative paragraph name, just realize that attenuation is inevitable but may not cause bias.
Since the parallel traditional medicine model is a special case of non-parallel targeted therapy model in which $\prescript{_Z}{}\beta^{Rx} = \prescript{_Z}{}\beta^{C}$, we use non-parallel targeted therapy model to illustrate attenuation in general.

Model (\ref{mod:DTZmodel}) which we take to be true have the predictor $Z$ measured \emph{without} error, so 
%add the content for parallel model?
\begin{eqnarray*}
\prescript{_Z}{}\beta^{Rx} = \frac{Cov(Y^{Rx[m]}_{r},Z_{r})}{Var({Z_r})} & , & \prescript{_Z}{}\beta^{C} = \frac{Cov(Y^{C[m]}_{s},Z_s)}{Var({Z_s})}.  
\end{eqnarray*}

% Let $(B_1,B_2,\ldots,B_p)$ be a vector of $ p $ biomarkers and let $ B = g(B_1,B_2,\ldots,B_p) $ be a biomarker \emph{score} calculated to be on the same scale as the response (the primary endpoint), with $B$ potentially predictive of outcome.  

% Suppose in practice, for patients $r$ and $s$, $B_r = Z_{r}+\prescript{_B}{}E^{[1]}_{r}$ and $B_s = Z_{s}+\prescript{_B}{}E^{[1]}_{s}$ with measurement errors are used instead of their (not observable) $ Z $ values for prediction and  ordinary regression is applied to the models 
% \begin{eqnarray}
% Y^{Rx[m]}_{r} & = & \prescript{_B}{}\alpha_{} + \prescript{_B}{}\gamma^{Rx} + \prescript{_B}{}\beta^{Rx}B_{r} + \prescript{_B}{}\epsilon^{Rx[m]}_{r} \nonumber\\
% Y^{C[m]}_{s} & = & \prescript{_B}{}\alpha_{} + \prescript{_B}{}\gamma^{C} + \prescript{_B}{}\beta^{C}B_{s} + \prescript{_B}{}\epsilon^{C[m]}_{s}.\label{mod:DTWmodel}
% % jch 21Mar2024 seems to me Z should be B = baseline
% \end{eqnarray}
% \noindent as if $B_r$ and $B_s$ were measured without errors.
% We show intuitively how attenuation occurs, causing potential bias.
We show intuitively how attenuation occurs.

Substituting $Z_{r} = B_{r} - \prescript{_B}{}E^{[1]}_{r}$ and $Z_{s} = B_{s} - \prescript{_B}{}E^{[1]}_{s}$ into (\ref{mod:DTZmodel}), we obtain
\begin{eqnarray*}
Y^{Rx[m]}_{r} & = & \prescript{_Z}{}\alpha+\prescript{_Z}{}\gamma^{Rx}+\prescript{_Z}{}\beta^{Rx} (B_{r} - \prescript{_B}{}E^{[1]}_{r}) + \prescript{_Z}{}\epsilon^{Rx[m]}_{r}\\
& = & \prescript{_Z}{}\alpha+\prescript{_Z}{}\gamma^{Rx}+\prescript{_Z}{}\beta^{Rx} B_{r} + \prescript{*}{}\epsilon^{Rx[m]}_{r} \\
Y^{C[m]}_{s} & = & \prescript{_Z}{}\alpha+\prescript{_Z}{}\gamma^{C}+\prescript{_Z}{}\beta^{C} (B_{s} - \prescript{_B}{}E^{[1]}_{s}) + \prescript{_Z}{}\epsilon^{C[m]}_{s}\\
& = & \prescript{_Z}{}\alpha+\prescript{_Z}{}\gamma^{C}+\prescript{_Z}{}\beta^{C} B_{s} + \prescript{*}{}\epsilon^{C[m]}_{s}
\end{eqnarray*}
where error $\prescript{*}{}\epsilon^{Rx[m]}_{r} = \prescript{_Z}{}\epsilon^{Rx[m]}_{r} - \prescript{_Z}{}\beta^{Rx} (\prescript{_B}{}E^{[1]}_{r})$ and $\prescript{*}{}\epsilon^{C[m]}_{s} = \prescript{_Z}{}\epsilon^{C[m]}_{s} - \prescript{_Z}{}\beta^{C} (\prescript{_B}{}E^{[1]}_{s})$.
% with error $\epsilon^{*} = (\epsilon - \beta_{C}^{baseline} E)$ and covariate $baseline$.
The two errors $\prescript{*}{}\epsilon^{Rx}_{r}$ and $\prescript{*}{}\epsilon^{C}_{s}$ are \textit{correlated} to the covariate $ B_{r} $ and $ B_{s} $ because they share the same component $ \prescript{_B}{}E^{[1]}_{r} $ and $ \prescript{_B}{}E^{[1]}_{s} $.

Take treatment $ C $ as an example and similarly for $ Rx $.
This correlation between error and the covariate causes attenuation because 
with $Z_{s}$ and $\prescript{_B}{}E^{[1]}_{s}$ being independent, 
\begin{eqnarray*}
Cov(Y^{Rx[m]}_{s}, B_{s}) & = & Cov(Y^{Rx[m]}_{s}, Z_{s} + \prescript{_B}{}E^{[1]}_{s})\\
& = & Cov(Y^{Rx[m]}_{s}, Z_{s})\\
Var(B_{s}) & = & Var(Z_{s} + \prescript{_B}{}E_{s}^{[1]})\\
& = & Var(Z_{s}) + Var(\prescript{_B}{}E^{[1]}_{s})
\end{eqnarray*}
with an extraneous $ Var(\prescript{_B}{}E^{[1]}_{s})$ for $Var(B_{s})$.  
Thus, comparing (\ref{mod:DTWmodel}) with (\ref{mod:DTZmodel}), we see $ \prescript{_B}{}\beta^{C} $ is \textit{attenuated} towards zero as emphasized in the measurement error literature (e.g., \cite{CarrollEtAl(2006)}, p.43), 
% $$ \beta_{Rx}^{B} = E [ \hat{\beta}_{Rx}^{B} ] = \frac{ Var(Z)}{ Var(Z)+Var(E)}\prescript{_Z}{}\beta^{Rx}.$$ 
$$ \prescript{_B}{}\beta^{C} = \mathbbm{E} [ \prescript{_B}{}{\hat{\beta}}^{C} ] = \frac{ Var(Z_{s})}{ Var(Z_{s})+Var(\prescript{_B}{}E^{[1]}_{s})}\prescript{_Z}{}\beta^{C},$$ 
similarly for $ \prescript{_B}{}\beta^{Rx} $.

The attenuation can be quantified by the ETZ modeling principle in one special case where $ B $ stands for \textit{baseline}, the primary endpoint itself at first visit.
In this case,
\begin{eqnarray}
B & = & Z + E^{[1]} \nonumber \\
~& = & Z + E, ~\mathrm{and} \nonumber \\
\prescript{_B}{}\beta^{C} & = & \mathbbm{E} [ \prescript{_B}{}{\hat{\beta}}^{C} ] \nonumber \\
& = & \frac{Var(Z)}{Var(Z)+Var(E)}\prescript{_Z}{}\beta^{C}, \label{eq:Attenuation}
\end{eqnarray}
noting the extra $ Var(E) $.  

We show below the presence of measurement error $ \prescript{_B}{}E $ or $ E $ in using \textit{baseline} or any other biomarker $ B $ score as a surrogate for the true patient effect $ Z $ causes \textit{attenuation} in estimating the regression coefficient of $ B $, which in turn can cause bias in estimating $Rx: C$ efficacy.
While future ETZ research might mitigate this bias, 
for now we show that estimated counterfactual efficacy is unbiased provided that efficacy is defined as ``on average''.
\subsection{Possible bias caused by attenuation}\label{sec:CFbiased}
Bias would occur if one predicts efficacy in subgroups of patients (e.g., Alzheimer's disease patients with \textit{mild} or \textit{severe} symptoms at baseline) when RHCE is modeled (without stratification) for the entire population or DTCE is trained on the entire population.  

%Basically, efficacy is defined off-centered for subgroup identification.
%During subgroup identification, patients are separated into different subgroups referring to biomarker values.
%For different biomarker values, the target is to find the optimal treatment with highest efficacy.
%%Put the plots here?
%%I think here is suitable to add some content to fix the mistake made in confidence band paper
These biases can be visualized.  
We use the outcome instrumental Abilities of Daily Living (iADL) for Alzheimer's disease to illustrate.  
Abilities to manage finances and to prepare food are some of the items measured by \textit{iADL}.

%introduce the model first and then the setting of parameters
% Hence, for every patient, we generate both potential outcome $ Rx $ and $ C $ together with one \textit{baseline} and one true first visit outcome without measurement error $ Z $.

We simulate potential $Rx$ and $C$ outcomes at milestone visit for 500 patients in a \textit{counterfactual} setting from the ETZ modeling principle with first visit, \textit{baseline} (before randomization) measurements 
\begin{eqnarray*}
Y^{[1]}_i & = &Y^{[1]}_i(Rx)\\
& = &Y^{[1]}_i(C)\\
& = &Z_i + E^{[1]}_{i},\ i = 1,\dots,500 
\end{eqnarray*}
and milestone visit measurements 
\begin{eqnarray}
Y^{[m]}_i(Rx) & = & Z_i + Traj^{Rx[m]}_{i} + E^{Rx[m]}_{i}, \\
Y^{[m]}_i(C) & = & Z_i + Traj^{C[m]}_{i} + E^{C[m]}_{i},\\
&~&i = 1,\dots,500.\nonumber
\end{eqnarray}
In ETZ modeling, the three components $Z$, $Traj$, and $E$ follow different Normal, independent, distributions. 
$Z_i$ is generated from a Normal distribution with mean $ \alpha $ and variance $Var(Z)$. 
$Traj^{Rx[m]}_{i}$ and $Traj^{C[m]}_{i}$ are generated from Normal distributions with means $\mu^{Rx}$ and $ \mu^{C} $ and common variance $Var(Traj)$, independently.  
$E^{[1]}_i$, $E^{Rx[m]}_{i}$ and $E^{C[m]}_{i}$ are generated from Normal distributions with mean 0 and variance $Var(E)$, independently.  
In the counterfactual setting, outcomes $ Rx $ and $ C $ for patient $ i $ share the same $Z_i$.  

Summary statistics from the EXPEDITION3 study and ETZ modeling principle in Table \ref{tab:EXPEDITION3-EliLilly} are used for data generation.  
\begin{table}[hbtp] 
               \centering
               \begin{tabular}{|l|r|r|}
                              \hline
                              & \multicolumn{1}{l|}{Variance} & \multicolumn{1}{l|}{SD} \\
                              \hline
                              Visit 1 (baseline, $Y^{[1]}$) & 64.58 & 8.04 \\
                              \hline
                              Visit 7 (milestone visit, $Y^{[m]}$) & 135.39 & 11.64\\
                              \hline
                              Change-from-baseline ($Y^{[change]}$)& 92.37 & 9.61 \\
                              \hline
                           \hline
                           Intercept ($Z$) & 53.80 & 7.34 \\
                            \hline
                            Trajectory ($Traj$) & 70.81 & 8.42 \\
                            \hline
                          Measurement Error ($E$)& 10.78 & 3.28 \\
                          \hline
               \end{tabular}%
               \label{tab:EXPEDITION3-EliLilly}%
               \caption{Variance and standard deviation (SD) of iADL at Visits 1 and $m$, and change-from-baseline of the EXPEDITION3 study are as reported in \cite{honig2018trial}. Vriances of the Intercept, Trajectory and Measurement Error are computed by the ETZ transformation.}
\end{table}%
% Construct a table to store the statistics obtained from EXPEDITION3
% Describe the distributions followed by the three components
%Assume that both potential outcome $ Rx $ and $ C $ for same patient $ i $ can be measured factually.
%$Z_i$ denotes the true first visit outcome and $ Y^{[m]}_i = B_i = Z_i + E^{[1]}_i$ is the first visit outcome measured.

\begin{figure}[hbtp]
\centering
\includegraphics[scale = 0.8]{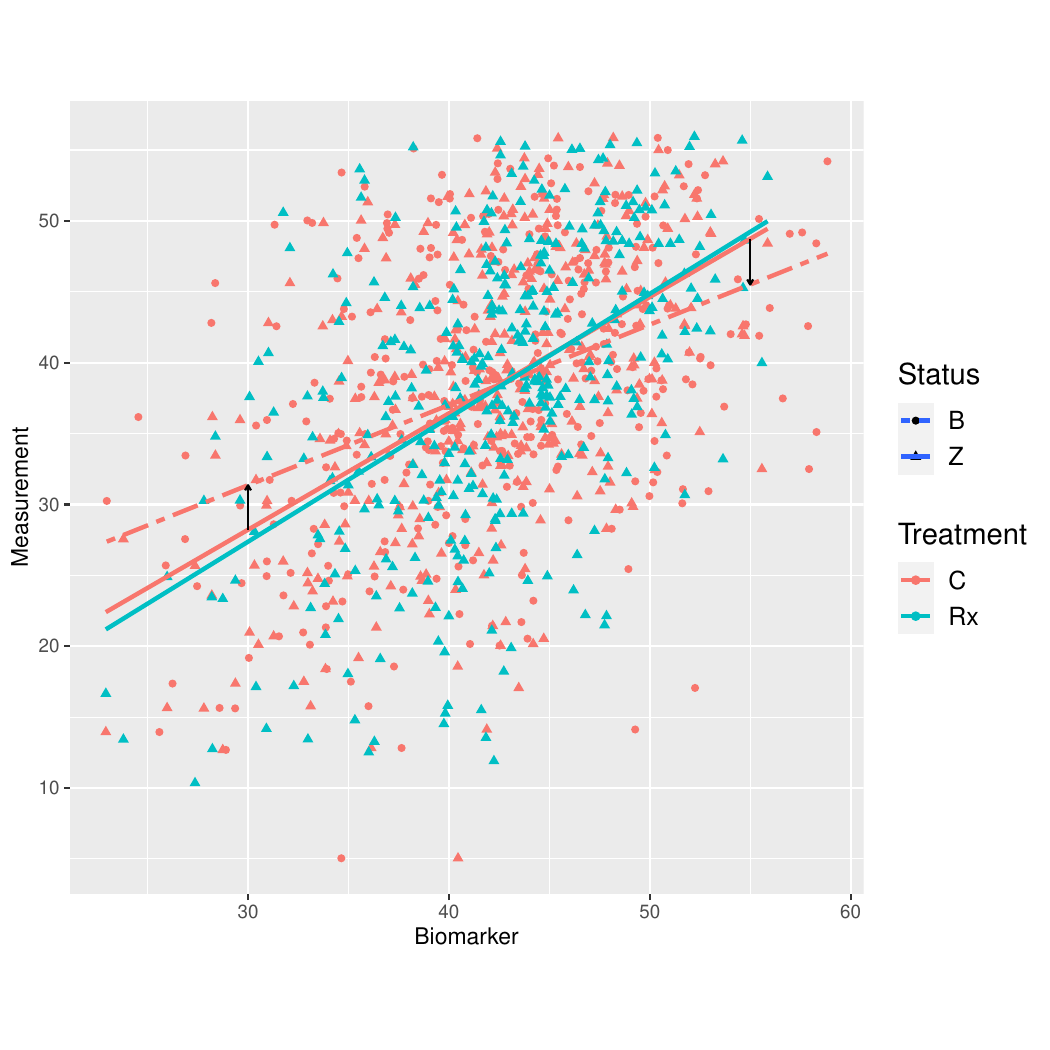}
\caption{This plot shows attenuation when $ B $ is a surrogate of the true $ Z $ measured with error. The slope regressing $ Y $ on $ B $ tends to be closer to zero, compared to the true slope regressing $ Y $ on $ Z $. Black arrows in the plot show attenuation bias in estimated $ C $ effect, due to measurement error in biomarker score $ B $ as a surrogate of the true $ Z $. Treatments are indicated by color while shape differentiates $ B $ from $ Z $. \textcolor{salmon}{Red} represents control \textcolor{salmon}{$ C $} while \textcolor{teal}{teal} represents treatment \textcolor{teal}{$ Rx $}. \textcolor{salmon}{\textbf{Red round points}} are $ C $ versus $ B $ measurements, with the \textcolor{salmon}{\textbf{red dashed line}} being the fitted $ C $ versus $ B $ regression line. \textcolor{salmon}{\textbf{red dashed line}} are $ C $ versus $ Z $ measurements, with the \textcolor{salmon}{\textbf{red solid line}} being the fitted $ C $ versus $ Z $ regression line. \textcolor{teal}{\textbf{Teal triangle points}} are $ Rx $ versus $ Z $  measurements, with the \textcolor{teal}{\textbf{teal solid line}} being the fitted $ Rx $ versus $ Z $ regression line.}
\label{fig:AttenuationPlot}
\end{figure}
\begin{figure}[hbtp]
\centering
\includegraphics[scale = 0.9]{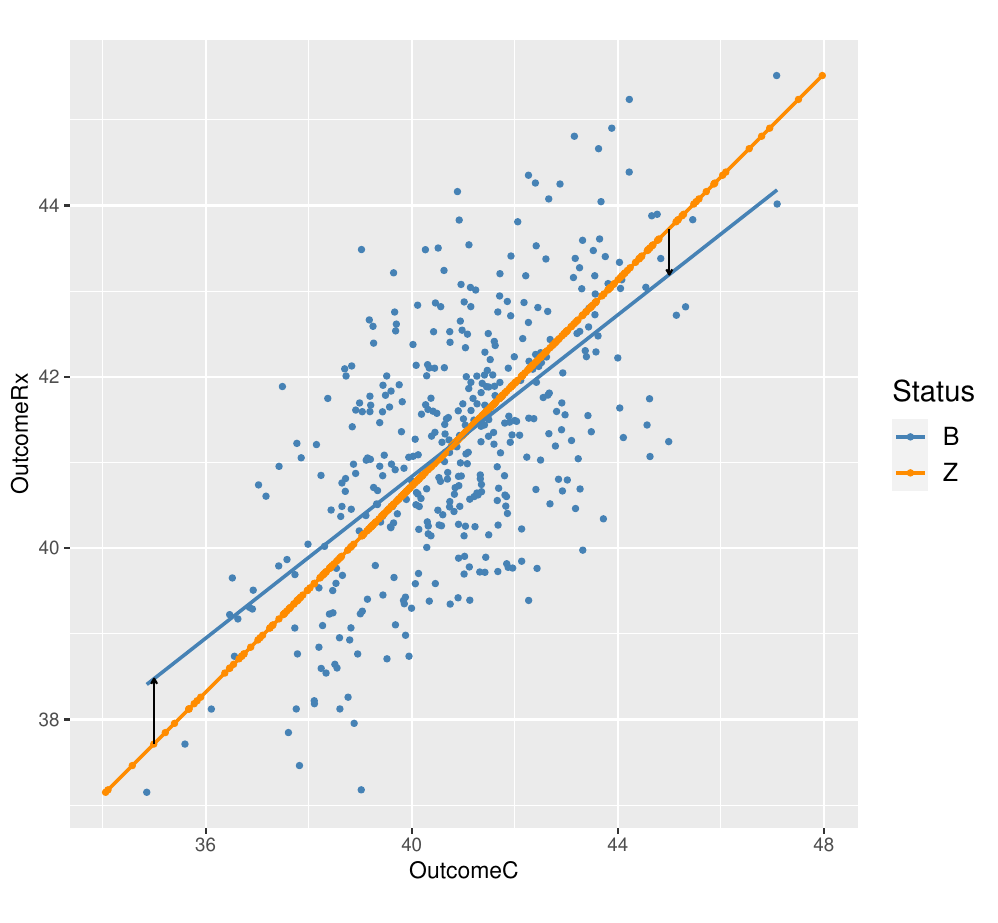}
\caption{This plot shows how bias in $ Rx $ versus $ C $ efficacy estimation depends on the target value of outcome $ C $.  
The \textcolor{darkorange}{\textbf{orange points}} represent outcome $ Rx $ (vertical axis) as predicted by true $ Z $ plotted against outcome $ C $ (horizontal axis) as predicted by true $ Z $, with the \textcolor{darkorange}{\textbf{orange line}} being the prediction line.  
The \textcolor{steelblue}{\textbf{blue points}} represent outcome $ Rx $ (vertical axis) as predicted by true $ Z $ plotted against outcome $ C $ (horizontal axis) as predicted by $ B (=Z+E) $, with the \textcolor{steelblue}{\textbf{blue line}} being their fitted line.  
$ Rx $ versus $ C $ efficacy, the average difference of response given $ Rx $ and response given $ C $, is biased upward when target $ C $ value is below average (with the \textcolor{steelblue}{\textbf{blue points}} trending above the \textcolor{darkorange}{\textbf{orange points}}), downward when target $ C $ value is above average (with \textcolor{steelblue}{\textbf{blue points}} trending below \textcolor{darkorange}{\textbf{orange points}}).  
}
% \caption{The direction of bias in $ Rx: C $ efficacy estimation depends on the target value of outcome $ C $. Efficacy is overestimated when target $ C $ value is below average, and underestimated when target $ C $ value is above average. For the \textbf{red points}, the (horizontal) abscissa (coordinate) is outcome $ C $ as predicted by $ B $, and the (vertical) ordinate (coordinate) is outcome $ Rx $ as predicted by $ Z $. The red line is the fitted line of $ Rx $ as predicted by true $ Z $ regressed on $ C $ as predicted by $ B $. For the \textbf{teal points}, both the abscissa outcome $ C $ and the ordinate outcome $ Rx $ are as predicted by true $ Z $. The teal line is the fitted line of $ Rx $ as predicted by true $ Z $ regressed on $ C $ as predicted by true $ Z $.  For easier visual perception, the $Rx$ and $C$ outcome values plotted have been partially denoised using values predicted by $ Z $ after we reduce the standard deviation of $ Z $ from the original 7.34 to 3.}
\label{fig:DigitalTwinPlot}
\end{figure}

Using \textit{baseline} or, more generally, a biomarker score $B$ to predict the outcome is tempting because $Z$ is not observable.  
However, as Figures \ref{fig:AttenuationPlot} and \ref{fig:DigitalTwinPlot} show, one should be aware that predicting the expected outcome at a $B$ value off-center may incur bias if $B$ has measurement error. 

\subsection{\texorpdfstring{Estimated average $Rx{:}C$ efficacy is unbiased}{Estimated average Rx:C efficacy is unbiased}}\label{sec:CFEUnbiased}
Population effect of treatment $Rx$ or $C$ is defined as response $Y$ averaged over infinitely many patients, each with their value $Z$.  
In the population space, what is of interest is the value $\mathbbm{E}[Y]$ at $ \mathbbm{E}[Z]$.

One intuitive way of seeing why the estimated average effect is unbiased is to imagine regressing $Y$ on $Z$ in (\ref{mod:DTZmodel}) versus regressing $Y$ on $B$ in (\ref{mod:DTWmodel}), with a rather large number of patients.  
Attenuation, which occurs when $B$ is a surrogate of $Z$ but measured with error, means the slope regressing $Y$ on $B$ in (\ref{mod:DTWmodel}) tends to be closer to zero, compared to the true slope regressing $Y$ on $Z$ in (\ref{mod:DTZmodel}).  

A familiar experience in simple linear regression is the fitted regression line goes through the central point $(\bar{X}, \bar{Y})$ of average predictor and response values.  
Thus, imagining the clinical study being repeated, one can visualize the pairs of regression lines from fitting (\ref{mod:DTZmodel}) and (\ref{mod:DTWmodel}) pivoting/rotating around the same central point, as depicted in Figure \ref{fig:AttenuationPlot}.  
Since $\mathbbm{E}[B] = \mathbbm{E}[Z]$, even though the two regression lines have different slopes, their values at $\mathbbm{E}[Z]$ (average value of $Z$) are the same.  

%Intuitively, that estimated average efficacy is unbiased can be explained from two perspectives.
%First, the regression lines described by (\ref{mod:DTWmodel}) and (\ref{mod:DTZmodel}), no matter whether there is a measurement error in biomarker or not, pass the same point in population space.
%The regression line described by (\ref{mod:DTWmodel}) is obtained by rotating (\ref{mod:DTZmodel}) around the certain point when there is attenuation.
%It is also verified by Figure \ref{fig:AttenuationPlot}.
%In population space, the intersection point is $ (\mathbbm{E}[Z],\mathbbm{E}[Y]) $.
%For infinity patients, the population average of biomarker with measurement error $\mathbbm{E}[B] = \mathbbm{E}[Z]$ because the expectation of measurement error $\mathbbm{E}[E] = 0$.
%Hence, when there is a sufficient number of patients in sample space, (\ref{mod:DTWmodel}) and (\ref{mod:DTZmodel}) will intersect at $(\Bar{B},\Bar{Y})$, where $\Bar{B}$ and $\Bar{Y}$ are the sample average of $B$ and $Y$ and approximate to $\mathbbm{E}[Z]$ and $\mathbbm{E}[Y]$ as number of patients goes infinity, respectively.

To explain unbiasedness mathematically, we 
define efficacy more generally as measured at $\mathbbm{E}[Z]\pm c$  where $ c $ is any non-negative constant.  
(Population efficacy as described earlier would be at $\mathbbm{E}[Z]$ with $c=0$.)  
We prove that:
\begin{itemize}
\item For traditional medicine, estimated \textit{equipoise} $Rx: C$ efficacy defined at $\mathbbm{E}[Z]\pm c$ is unbiased for all $ c \ge 0$.  
\item For targeted therapy, \textit{control-side} and \textit{equipoise} $Rx: C$ efficacy estimates are only unbiased when $c = 0$, but can be biased when efficacy $\mathbbm{E}[Z]\pm c$ is defined off-centered with $ c > 0 $. 
\end{itemize} 

Traditional medicine has parallel profiles.  
First, we caution \textit{control-side} $Rx: C$ efficacy estimation is biased and should not be done, for the following reason.  
For traditional medicine, treatment $ Rx $ and treatment $ C $ share the same coefficient for $ baseline $ in both model (\ref{mod:ParallelmodelZ}) and (\ref{mod:ParallelmodelB}).
For \textit{control-side} estimation, counterfactual $ C $ outcomes are predicted through model (\ref{mod:ParallelmodelB}) with attenuated \textit{baseline} coefficient $ \prescript{_B}{}\beta = \frac{Var(Z)}{Var(Z)+Var(\prescript{_B}{}E)}\prescript{_Z}{}\beta $, while factual $ Rx $ outcomes follow the true model (\ref{mod:ParallelmodelZ}) with coefficient of \textit{baseline}, $ \prescript{_Z}{}\beta $.
Therefore, as shown in Figure \ref{fig:TradTargetCom}, control-side estimation makes the apparent profiles not parallel for traditional medicine and should not be done.  

Now we prove \textit{equipoise} $Rx: C$ efficacy estimation is unbiased for traditional medicine mathematically, no matter what value $c \ge 0$ efficacy  $\mathbbm{E}[Z]\pm c$ is defined at.  
For \textit{equipoise} counterfactual estimation, efficacy is measured within each of $ Rx $ and $ C $ first and then combined.
Efficacy is measured as the difference between a factual measured outcome and its corresponding \textit{counterfactual} outcome within each treatment arm which, 
assuming the numbers of patients in $ Rx $ and $ C $ are equal, is 
\begin{small}
\begin{align*}
	&\frac{1}{2}[(\prescript{_Z}{}\theta_{} + \prescript{_Z}{}\eta^{Rx} + \prescript{_Z}{}\beta(\mathbbm{E}[Z]\pm c))\\\nonumber
	&-(\prescript{_B}{}\theta_{} + \prescript{_B}{}\eta^{C} + \prescript{_B}{}\beta(\mathbbm{E}[Z]\pm c))]\\\nonumber
	&+\frac{1}{2}[(\prescript{_B}{}\theta_{} + \prescript{_B}{}\eta^{Rx} + \prescript{_B}{}\beta(\mathbbm{E}[Z]\pm c))\\\nonumber
	&-(\prescript{_Z}{}\theta_{} + \prescript{_Z}{}\eta^{C} + \prescript{_Z}{}\beta(\mathbbm{E}[Z]\pm c))]\\\nonumber
	&= \mathbbm{E}[Y^{Rx[m]}_{r}]-\mathbbm{E}[Y^{C[m]}_{s}].
\end{align*}
\end{small}
Hence, \textit{equipoise} counterfactual estimation is unbiased for traditional medicine.

We discuss both \textit{control-side} and \textit{equipoise} counterfactual estimations for targeted therapy.
For \textit{control-side} counterfactual estimation of targeted therapy, efficacy is measured as the difference between the factual outcome of a patient receiving $ Rx $ and the corresponding \textit{counterfactual} $ C $ outcome: 
\begin{eqnarray*}
&~&(\prescript{_Z}{}\alpha_{} + \prescript{_Z}{}\gamma^{Rx} + \prescript{_Z}{}\beta^{Rx}(\mathbbm{E}[Z]\pm c))-\\
&~&(\prescript{_B}{}\alpha_{} + \prescript{_B}{}\gamma^{C} + \prescript{_B}{}\beta^{C}(\mathbbm{E}[Z]\pm c))\\\nonumber
&=& (\mathbbm{E}[Y^{Rx[m]}_{r}]-\mathbbm{E}[Y^{C[m]}_{s}])\pm\\    
&~&c(\prescript{_Z}{}\beta^{Rx}-\frac{Var(Z)}{Var(Z)+Var(\prescript{_B}{}E)}\prescript{_Z}{}\beta^{C})\nonumber
\end{eqnarray*}
so there is bias if $c >0$.  

\textit{Equipoise} counterfactual estimation for efficacy in targeted therapy can be expressed as:
% so that the efficacy is mixed equally.
\begin{small}
\begin{align*}
	&\frac{1}{2}[(\prescript{_Z}{}\alpha_{} + \prescript{_Z}{}\gamma^{Rx} + \prescript{_Z}{}\beta^{Rx}(\mathbbm{E}[Z]\pm c))\\\nonumber
	&-(\prescript{_B}{}\alpha_{} + \prescript{_B}{}\gamma^{C} + \prescript{_B}{}\beta^{C}(\mathbbm{E}[Z]\pm c))]\\\nonumber
	&+\frac{1}{2}[(\prescript{_B}{}\alpha_{} + \prescript{_B}{}\gamma^{Rx}+ \prescript{_B}{}\beta^{Rx}(\mathbbm{E}[Z]\pm c))\\\nonumber
	&-(\prescript{_Z}{}\alpha_{} + \prescript{_Z}{}\gamma^{C} + \prescript{_Z}{}\beta^{C}(\mathbbm{E}[Z]\pm c))]\\\nonumber
	&= (\mathbbm{E}[Y^{Rx[m]}_{r}]-\mathbbm{E}[Y^{C[m]}_{s}])\\
	&\pm\frac{c}{2}(\prescript{_Z}{}\beta^{Rx}-\frac{ Var(Z)}{ Var(Z)+Var(\prescript{_B}{}E)}\prescript{_Z}{}\beta^{C})\\
	&\pm\frac{c}{2}(\frac{ Var(Z)}{ Var(Z)+Var(\prescript{_B}{}E)}\prescript{_Z}{}\beta^{Rx}-\prescript{_Z}{}\beta^{C})\nonumber
\end{align*}
\end{small}
so there is bias if $c >0$.  
That would happen if the target patient population in the actual RCT does not align with the patient population in the learning study.  
Table \ref{tab:MMSE} advises extraordinary caution in this situation using two examples.  	

    \begin{table}[hbtp] \centering 
		\begin{tabular}{|c|c c|}
			\hline
			& \multicolumn{2}{c|}{MMSE scoring} \\
			% \hline
			Study & Mild/Early & Moderate \\
			\hline \hline
			EXPEDITION1 & $\checkmark$ & $\checkmark$ \\
			\hline
			Start of EXPEDITION2 & $\checkmark$ & $\checkmark$ \\
			\hline
			End of EXPEDITION2 & $\checkmark$ & \\
			\hline
			EXPEDITION3 & $\checkmark$ & \\
			\hline \hline
			Training ProCovA & $\checkmark$ & \\
			\hline
			Applying ProCovA & $\checkmark$ & $\checkmark$ \\
			\hline
		\end{tabular}
		\caption{From Section \ref{sec:CFbiased}, and experience from EXPEDITION1/2/3 (which did not meet their primary objectives), extraordinary caution is advisable in predicting potential outcome when the predictor contains measurement error and the learning study patient population differs from the application study patient population.  
		EXPEDITIONS 1, 2, and 3 staged Alzheimer's disease patients by the  Mini–Mental State Examination (MMSE) score, with an MMSE score of 20 to 26 classified as \textit{mild}, while a score of 16 to 19 is \textit{moderate}.  
		From \cite{honig2018trial}, MMSE has an estimated $Var(E)$ of 2.06 which is not ignorable since the estimated MMSE difference between solanezumab and the placebo is 0.49.  EXPEDITION 2 started by targeting \textit{mild} and \textit{moderate} patients, but changed to targeting \textit{mild} patients after learning from EXPEDITION1’s result.  EXPEDITION3 targeted only \textit{mild} patients after learning from data on \textit{mild} patients pooled from EXPEDITION1 and EXPEDITION2.  From \cite{schuler2022increasing}, it appears ProCovA was applied to \textit{mild} and \textit{moderate} patients with MMSE between 14 and 26 in the DHA supplement study, while the algorithm appears to have been trained on \textit{early stage} patients.}\label{tab:MMSE}
	\end{table}

\section{Guidelines for applying the ETZ modeling principle}

ETZ is a modeling \emph{principle} that enables Counterfactual Uncertainty Quantification (CUQ) for randomized studies with Before-and-After Repeated Measures (BAtRM).  
Scientists faced with highly variable outcome measure (i.e., large measurement error) situations can benefit from this principle, provided that they can extract the three inputs needed by ETZ, working out details for proper CUQ in their settings themselves.  

For example, highly variable responses to therapies for amyotrophic lateral sclerosis (ALS, affecting $\approx$ 31,000 patients in the U.S.), measured on the ALS Functional Rating Scale–Revised (ALSFRS-R) scale (\cite{AMX003FDAstatisticalReview(2021)}) can be modeled with a random coefficient BAtRM model (e.g., using the RANDOM statement in Proc Mixed of SAS), providing inputs needed by ETZ.  
As another example, RCTs in immunology such as for systemic lupus erythematosus (SLE, affecting $\approx$ 200,000 patients in the U.S.), often are BAtRM studies with binary endpoints.  
% jch 12Apr2025
%For example, Morand et al.\ (2023) reported result of the Phase 3 study SLE-BRAVE-I, testing baricitinib for the treatment of systemic lupus erythematosus (SLE). 
%SLE-BRAVE-I was a double-blind, randomized, placebo-controlled, parallel-group Phase 3 study. 
%A total of 760 participants were randomly assigned to receive at least one dose of baricitinib 4 mg (n=252), baricitinib 2 mg (n=255), or placebo (n=253).  
%Primary endpoint was the proportion of patients achieving an SLE Responder Index (SRI)-4 response at week 52 in the baricitinib 4 mg treatment group compared to the placebo group. Achieving  SRI-4 is a binary outcome, a composite responder index that includes improvement in disease activity (defined as at least a 4-point improvement in the SLEDAI-2K score) without worsening of the overall condition (no increase of $\ge$ 0.3 points [10 mm] in the Physician Global Assessment from baseline) or the development of significant disease activity in new organ systems (no new BILAG A or more than one new BILAG B).
%
% jch 12Apr2025
%To develop treatments for diseases such as ALS (which affects an estimated 31,000 patients in the U.S.) and SLE (which affects approximately 200,000 patients in the U.S.), variability in the outcome measure is a known challenge.  
For binary and time-to-event outcomes, efficacy is often measured as a ratio.  
The CUQ principle for reducing uncertainty quantification can be applied to ratio efficacy, provided one is aware of the following: 
\begin{enumerate}
	\item Mixing not-logic-respecting efficacy measures Odds Ratio (OR) and Hazard Ratio (HR) in subgroups \textit{attenuates} them, even without measurement error in (biomarker) predictors, see \cite{LiuWangKilHsu(2022)}; 
	\item Computer packages currently give misleading stratified analysis results for all ratio efficacy (logic-respecting or not), because the \emph{prognostic} effect is ignored, see \cite{LiuEtAl(2023)}.
\end{enumerate}
However, following the Subgroup Mixable Estimation (SME) principle, correct causal inference for binary and time-to-event outcomes can be obtained for logic-respecting rato efficacy such as Relative Response (RR) and Ratio-of-Time, see \cite{Ding&Lin&Hsu(2016)}, \cite{Lin&Xu&Ding&Hsu(2019)}, the discussion paper \cite{LiuWangKilHsu(2022)}, and its rejoinder \cite{LiuWangHongHsu(2022)}.    

% To develop treatments for diseases such as ALS (which affects an estimated 31,000 patients in the U.S.) and SLE (which affects approximately 200,000 patients in the U.S.), variability in the outcome measure is a known challenge.  
% The CUQ principle for reducing uncertainty quantification is developed to help meet this challenge, for the greater good.  

\section{Key messages}
\begin{tcolorbox}
\begin{itemize}
	\item ETZ is an uncertainty quantification (UQ)  \textit{principle} for Randomized Controlled Trials (RCTs) with \textit{Before-and-After} treatment Repeated Measures (BAtRM).
	\item Conducting RCT with BAtRM is standard practice in therapeutic areas affecting more than half a billion patients worldwide.  
	\item By modeling individual patient outcomes in BAtRM studies, ETZ enables \textit{counterfactual} uncertainty quantification (CUQ).
	\item CUQ is typically smaller than current \textit{factual} uncertainty quantification.  
	% \item Reduction in uncertainty quantification translates to reduction in sample size.
	\item Statistical or digital \textit{counterfactual} \textit{point} estimation involving a biomarker with measurement error is unbiased for efficacy in the entire population but \emph{biased} in subgroups.
\end{itemize}
\end{tcolorbox}

\section{Acknowledgments}

The ETZ modeling thinking started with Haiyan Xu saying “Dr. Hsu, you only need three numbers!”  

We thank the AE for on-point comments, particularly for requesting us to comment on ProCovA.

\section{Conflicts of interest}

X.W., Y.H., S-Y.T. declare no competing interests.\\
Y. L. is an employee and shareholder of Eli Lilly.\\
J.C.H. is a professor emeritus of the Ohio State University and a consultant to Eli Lilly.

\begin{appendix}
	\section{Glossary of terminologies}\label{appB}
	\begin{tcolorbox}
	%\begin{glossarybox}
		% Glossary entries
        \textbf{BAtRM}: \textit{Before-and-After} treatment Repeated Measures.\\
		\textbf{Effect} of a treatment: Expected outcome given treatment $ Rx $ or $ C $.\\  
		\textbf{Efficacy}: Expected differential outcome between treatments $ Rx $ and $ C $.\\  
		\textbf{Attenuation}: Directional bias (\textit{within} each treatment arm) in estimating slope of the treatment effect when predictor contains measurement error.\\
		\textbf{Factual study}: A study in which each patient is given either $Rx$ \textbf{or} $C$.\\ %(e.g., studies with parallel designs).\\
		\textbf{\textit{Counterfactual} study}: A study in which each patient is given both $Rx$ \textbf{and} $C$.\\ % (e.g., a crossover study).\\
		\textbf{\textit{Counterfactual} efficacy}: Expected differential between $Rx$ and $C$ outcomes of patients in a \textit{counterfactual} study.\\
		\textbf{\textit{Counterfactual} estimation (CE)}: Unbiased \textit{point} estimation of \textit{counterfactual} efficacy from a \emph{factual} study.\\
		\textbf{\textit{Counterfactual} uncertainty quantification (CUQ)}: Quantification of uncertainty (such as variance) of a \textit{counterfactual} point estimate calculated as if it were from a \textit{counterfactual} study.\\
		\textbf{Targeted therapies}: Therapies such as tyrosine-kinase inhibitors and monoclonal antibodies that target specific pathways, constituting a large portion of ``personalized'' or ``precision'' medicine.\\  
		\textbf{Real Human \textit{Counterfactual} Estimation (RHCE)}: Counterfactual point estimation (CE) with Counterfactual Uncertainty Quantification (CUQ) using each real human patient as self-control.\\
		\textbf{Digital Twins \textit{Counterfactual} Estimation (DTCE)}: \textit{Counterfactual} point estimation (CE) using biomarkers to predict for each $ Rx $ patient its digital $ C $ twin's outcome and vice versa, then averaging the outcome differentials.   
%	\end{glossarybox}
\end{tcolorbox}

	%\section{Notation}\label{appC}
	
	% \begin{glossarybox}
		%     % Notation entries
		%     $ Y_i (Rx)\textbackslash Y_i (C) $: The potential outcomes of $ Rx $ or $ C $ for patient $ i $.\\
		%     $ $

		% \end{glossarybox}
\section{Details of CUQ variance reduction calculations}\label{appC}
Detailed calculations of the two-step CUQ process that led to the pie chart for EXPEDITION3 in Figure \ref{fig:CUQexamples} are given here.  
Calculations for the AbbVie data are similar.

Table~\ref{tab:EXPEDITION3-EliLilly} presents summary statistics for EXPEDITION3 along with their ETZ variance components.  
%     \begin{table}[tbh] 
%                \centering
%                \begin{tabular}{|l|r|r|}
%                               \hline
%                               & \multicolumn{1}{l|}{Variance} & \multicolumn{1}{l|}{SD} \\
%                               \hline
%                               Visit 1 (baseline, $Y^{[1]}$) & 64.58 & 8.04 \\
%                               \hline
%                               Visit 7 (milestone visit, $Y^{[m]}$) & 135.39 & 11.64\\
%                               \hline
%                               Change-from-baseline ($Y^{[change]}$)& 92.37 & 9.61 \\
%                               \hline
%                            \hline
%                            Intercept ($Z$) & 53.80 & 7.34 \\
%                             \hline
%                             Trajectory ($Traj$) & 70.81 & 8.42 \\
%                             \hline
%                           Measurement Error ($E$)& 10.78 & 3.28 \\
%                           \hline
%                \end{tabular}%
%                \label{tab:EXPEDITION3-EliLilly}%
%                \caption{Variance and standard deviation (SD) of iADL at Visits 1 and $m$, and change-from-baseline of the EXPEDITION3 study are as reported in \cite{honig2018trial}. Vriances of the Intercept, Trajectory and Measurement Error are computed by the ETZ transformation.}
% \end{table}%
Detailed calculation showing baselining reduces variance in milestone visit by 43.02 which is 31.77\% of total variance is as follows.  
\begin{eqnarray*}
    Var (Y^{[m]}) & = & Var (Z) + Var (Trt) + Var (E)\\\nonumber
     & = & 135.39\\
    Var (Y^{[change]}) & = & Var (Trt) + 2\times Var (E)\\\nonumber
     & = & 92.37\\
    Var (Y^{[m]}) - Var (Y^{[change]}) & = & Var (Z) - Var (E) \\\nonumber
     & = & 53.802 - 10.778\\\nonumber
     & = & 43.02\\
    \frac{Var (Y^{[m]}) - Var (Y^{[change]})}{Var (Y^{[m]})} & = & 31.77\%.
\end{eqnarray*}
Detailed calculation showing the counterfactual aspect of CUQ further reduces variance by 21.56 beyond baselining by 11.67\% is as follows.
\begin{eqnarray*}
    Var (Y^{f}) & = & 2\times Var (Trt) + 4 \times Var (E)\\\nonumber
     & = & 2\times 70.809 + 4 \times 10.778\\\nonumber
     & = & 184.73\\
    Var (Y^{cf}) & = & 2\times Var (Trt) + 2 \times Var (E)\\\nonumber
    & = & 2\times 70.809 + 2 \times 10.778\\\nonumber
    & = & 163.174\\
    Var (Y^{f}) - Var (Y^{cf}) & = & 2 \times Var (E)\\\nonumber
    & = & 184.73 - 163.174\\\nonumber
    & = & 21.556\\
    \frac{Var (Y^{f}) - Var (Y^{cf})}{Var (Y^{f})} & = & 11.67\%.
\end{eqnarray*}
Hence, CUQ reduces
\begin{equation}
    31.77\%+11.67\%\times(1-31.77\%) = 39.73\%
\end{equation}
of the total variance in EXPEDITION3.
\end{appendix}

\bibliographystyle{Chicago}

\bibliography{Xreference}
\end{document}